%% file: BPSVRG.tex
\documentclass{article}

\usepackage[final]{neurips_2019}

\usepackage[utf8]{inputenc} % allow utf-8 input
\usepackage[T1]{fontenc}    % use 8-bit T1 fonts
\usepackage{hyperref}       % hyperlinks
\usepackage{url}            % simple URL typesetting
\usepackage{booktabs}       % professional-quality tables
\usepackage{amsfonts}       % blackboard math symbols
\usepackage{nicefrac}       % compact symbols for 1/2, etc.
\usepackage{microtype}      % microtypography
\usepackage{mathrsfs}
\usepackage{multicol}
\usepackage{algorithm}
\usepackage{algorithmic}
\usepackage{adjustbox}
\usepackage{color}
\usepackage{mathrsfs}
\usepackage{amsmath} %数学公式
\usepackage[numbers]{natbib}
\usepackage{float}
\usepackage{diagbox}
\usepackage{subcaption}
\usepackage{amssymb}
\usepackage[dvipsnames]{xcolor}
\usepackage{enumitem}
\newtheorem{thm}{Theorem}

\newtheorem{assume}[thm]{Assumption}

\title{Towards Better Generalization: \\BP-SVRG in Training Deep Neural Networks}

\author{%
  \textbf{Hao Jin} \\
  Peking University\\
  \texttt{jin.hao@pku.edu.cn} \\
  \and
  \textbf{Dachao Lin} \\
  Peking University\\
  \texttt{lindachao@pku.edu.cn}\\
  \And
  Zhihua Zhang \\
  Peking University \\
  Beijing, China \\
  \texttt{zhzhang@math.pku.edu.cn}
}

\begin{document}

\maketitle

\begin{abstract}
Stochastic variance-reduced gradient (SVRG) is a classical optimization method. 
Although it is theoretically proved to have better convergence performance than stochastic gradient descent (SGD), the generalization performance of SVRG remains open.
In this paper we investigate the effects of some training techniques, mini-batching and learning rate decay, on the generalization performance of SVRG,
% Such variants of SVRG with mini-batching techniques and learning rate decay is referred to as B-SVRG.
% To derive a practical variant of SVRG for deep neural network training, we firstly investigate , 
%  derive Batch-SVRG (B-SVRG) accordingly.
and verify the generalization performance of  Batch-SVRG (B-SVRG).
In terms of the relationship between optimization and generalization, we believe that the average norm of gradients on each training sample as well as the norm of average gradient indicate how flat the landscape is and how well the model generalizes.
Based on empirical observations of such metrics, we perform a sign switch on B-SVRG and derive a practical algorithm, \textit{BatchPlus-SVRG (BP-SVRG)}, which is numerically shown to enjoy better generalization performance than B-SVRG, even SGD in some scenarios of deep neural networks.

\end{abstract}

\input{intro}
\input{algo}
\input{motivation}
\input{generalization}
\input{exper}
\input{conclusion}

\newpage
\bibliography{BPSVRG}
\bibliographystyle{plainnat}

\newpage
\input{appendix}
\end{document}

%% file: intro.tex
\section{Introduction}
Deep neural networks have recently brought about great empirical improvements in various application tasks.
Training a deep neural network is tough, because the network is generally nonconvex with great complexity and overparameterization \cite{arora2018optimization}.
In spite of the theoretical toughness, many simple optimization methods are empirically proved to find solutions that generalize well beyond the training dataset \cite{neyshabur2015norm,neyshabur2015data}.
These optimization methods have been thoroughly analyzed when the objective function is convex or nonconvex with known Lipschitz continuity of the gradient \cite{schmidt2017minimizing,defazio2014saga,reddi2016stochastic,allen2016improved,allen2017katyusha}.
Therefore, it is natural to expect an optimization method with better theoretical result to enjoy better empirical performance, especially in terms of generalization.
However, the heartbroken fact is that a faster theoretical convergence rate is not a guarantee of better empirical performance in deep neural networks, just as it does in the case of stochastic variance reduced gradient (SVRG) \cite{defazio2018ineffectiveness}.

% Stochastic variance reduced gradient (SVRG) \cite{johnson2013accelerating} is one of the most popular optimization methods.
The training of deep neural networks in many tasks of machine learning can be extracted as finding an approximation solution to the following optimization problem
% in training neural networks is formulated as follows:
% \vspace{-0.2in}
\begin{equation*}
    % \notag
    \min_{w}~F(w)\triangleq \frac{1}{n}\sum_{i=1}^n f_i(w), \qquad \qquad
\end{equation*}
% \vspace{-0.15in}
where $w\in \mathbb{R}^d$ is the parameter vector, $n$ is the size of training dataset, and $f_i:\mathbb{R}^d\rightarrow \mathbb{R}$ indicates the specific loss function for the $i$-th training sample. Due to tremendous data size $n$ and some complex loss function (e.g. neural network) with highly nonconvex and nonsmooth loss landscape \cite{goodfellow2016deep}, stochastic gradient methods are widely used to leverage the sample efficiency and the provable convergence guarantee.  
The simplest stochastic optimization method, stochastic gradient descent (SGD), updates the parameters every time with a random sample $i_t$ chosen from the training dataset:
\begin{equation}
     w_{t{+}1} = w_{t}-\eta\nabla f_{i_t}(w_{t}).\qquad \qquad
\end{equation}
In theoretical analysis of optimization methods, the uncertainty introduced by the variance of stochastic gradient deteriorates the convergence speed of gradient descent (GD) from $O(\frac{n}{\epsilon})$ \cite{nesterov2013introductory, ghadimi2016accelerated} to $O(\frac{1}{\epsilon^2})$ \cite{nesterov2013introductory, reddi2016stochastic}.
In order to address the issue, SVRG somehow cancels out the variance of $\nabla f_{i_t}(w_t)$ with a control variate $\mu_{i_t}(w^{(s)})$ which has zero expectation:
\begin{equation}\label{svrg}
\begin{aligned}
&\mu_{i_t}(w^{(s)})=\nabla f_{i_t}(w^{(s)})-\nabla F(w^{(s)}),\\
&w_{t+1}=w_t-\eta[\nabla f_{i_t}(w_t)-\mu_{i_t}(w^{(s)})],
\end{aligned}
\end{equation}
where $w^{(s)}$ represents the snapshot point maintained by SVRG.
Specifically, SVRG estimates the gradient of entire objective function $\nabla F(w)$ for every snapshot point $w^{(s)}$, and $w^{(s)}$ is updated every $m$ steps of parameter update described in Eq.~\eqref{svrg}.
For a convex objective function, to obtain a $w$  such that $\|\nabla F(w)\|\leq \epsilon$, SVRG improves the times of gradient computation from $O(\frac{1}{\epsilon^2})$ of SGD to $O(n+\frac{\sqrt{n}}{\epsilon})$.
Such advantages of SVRG remain even when the objective function is nonconvex \cite{reddi2016stochastic}.

Different from SGD, the direct application of SVRG to deep neural networks is improper.
That is mainly due to the introduction of snapshot point $w^{(s)}$.
Specifically, the update frequency $m$ of the snapshot point $w^{(s)}$ is a hyper-parameter requiring special attention for the following reasons:
\begin{itemize}
\vspace{-0.05in}
    \item Each snapshot update requires a calculation of the full gradient $\nabla F(w)$.
    Considering the complexity of network structure and the enormous size of the training data, full gradient estimation is both time consuming and memory consuming.
    \item The distance between the current point $w_t$ and the later snapshot point $w^{(s)}$ should not be so long.
    When the current point $w_t$ goes further from the snapshot point $w^{(s)}$, the control variate $\mu_{i_t}(w^{(s)})$ comes increasingly uncorrelated with the current stochastic gradient $\nabla f_{i_t}(w_t)$, which is empirically shown to ruin the generalization performance \cite{defazio2018ineffectiveness}.
\end{itemize}
\vspace{-0.05in}
In this way, larger $m$ leads to less frequent calculation of $\nabla F(w)$, but increases the possibility of $\mu_{i_t}(w^{(s)})$ to be out of date; smaller $m$ guarantees the correlation between $\mu_{i_t}(w^{(s)})$ and $\nabla f_{i_t}(w_t)$, but increases the update frequency of the snapshot.
To address the dilemma, \cite{lei2017non} devised a variant of SVRG, which utilizes the mini-batching techniques in both inner iterations and outer iterations.
It is empirically turned out that mini-batching techniques with fixed batch size deteriorates the generalization performance of the original SVRG.
Learning rate decay manages to fulfill such generalization gap, deriving a practical variant of SVRG for deep neural networks.
For simplicity, we refer to such a SVRG variant as Batch-SVRG (B-SVRG)
(see Algorithm \ref{bsvrg}).

In this paper, we introduce the average norm of gradients on each training sample and the norm of  average gradient to measure the current generalization ability of a model. 
The sharpness of the objective function \cite{keskar2016large,neyshabur2017exploring} is a popular choice to explain the generalization performance: the flatter the landscape is, the better generalization performance the model enjoys.
After replacing the distribution of direction vector with a data relevant distribution, we discover that the average norm of gradients depicts the expected sharpness along the marching direction of a model.
Additionally, the norm of average gradient and the average norm of gradients are found to form an upper bound for the generalization gap when the objective functions satisfy the P-L conditions \cite{polyak1963gradient}.
Considering the great capacity of deep neural networks, the training loss is generally small through common optimization methods, and the generalization gap actually determines how well the model generalizes.

We make a sign switch on B-SVRG and derive a practical optimization method that we call \textit{BatchPlus-SVRG (BP-SVRG)} (see Algorithm \ref{bsvrg}).
Such a sign switch is motivated by the empirical observation on common optimization methods as well as B-SVRG.
Specifically, in deep neural networks, B-SVRG has larger values of the two metrics mentioned above, which partially explains its generalization performance.
The sign switch manages to control the norm of average gradient and the average norm of gradients during the training, and is empirically shown to help BP-SVRG obtain better generalization performance than SGD in common scenarios.
Actually, the derivation of BP-SVRG follows the trend of increasing the variance of stochastic gradient, just as it does in the case of larger learning rate and smaller batch size for SGD \cite{keskar2016large}.

%% file: algo.tex
\section{Training techniques on SVRG}
To find practical SVRG variants for deep neural networks, we consider the commonly used training techniques for optimization methods.
Batch-SVRG (B-SVRG) is a detailed implementation of algorithms in \cite{lei2017non}.
Specifically, B-SVRG fixes the batch size for inner iterations and outer iterations, and decays the learning rate throughout the training process.
In this section, we mainly focus on the generalization performance in deep neural networks, and challenge the models of AlexNet \cite{krizhevsky2012imagenet} on classification task of CIFAR10.
To reduce the randomness, experiments are repeated for three times for each setting.
Specifically, we discuss the effects of mini-batching on generalization performance of SVRG variants, and argue that learning rate decay is necessary for such SVRG variants to enhance the generalization performance.
In this way, we verify the comparable generalization performance of B-SVRG with the original SVRG in scenarios of deep neural networks.

\subsection{Mini-batching on SVRG}
Mini-batching techniques are applied to SVRG in both inner iterations and outer iterations.
Compared with the original SVRG, B-SVRG uses batches of data in starred lines (*) and (**).
The outer batch $I$ is selected from the training dataset $\mathcal{S}$ for the full gradient estimation at snapshot points, while the inner batch $\tilde{I}$ is selected from $I$ without replacement for the current stochastic gradient calculation.
The outer batch is designed to free the complexity of snapshot update from the total size of training data $\mathcal{S}$; the inner batch is designed to decrease the times of gradient computation without sacrificing much performance.
In terms of the special cases of Batch-SVRG, those with $B=|\mathcal{S}|$ are kind of the original SVRG, while those with $B=b$ are actually the SGD with a batch size of $b$.

\begin{algorithm}
    \begin{algorithmic}
    \caption{\colorbox{SpringGreen}{B-SVRG} and \colorbox{BurntOrange}{BP-SVRG}}
    \label{bsvrg}
    \STATE $\textbf{Input}$ Training dataset $\mathcal{S}$, learning rate $\{\eta_s\}$, outer batch size $B$, and inner batch size $b$ ($B\geq b$), number of epochs $T$.
    \STATE $\textbf{Initialize}$ $w^{(0)}$
    \STATE \textbf{for} $s=1,~2,~\ldots,~T$ \textbf{do}
    \STATE \quad $w=w^{(s{-}1)}$
    \STATE \quad $I$ randomly selected from $\mathcal{S}$ with $|I|=B$ \hfill (*)
    \STATE \quad $\mu=\frac{1}{B}\sum_{i\in I}\nabla f_i(w)$
    \STATE \quad $w_0=w$
    \STATE \quad \textbf{for} $t=1,~2,~\ldots,~\lfloor\frac{B}{b}\rfloor$ \textbf{do}
    \STATE \quad \quad $\tilde{I}=I[(t-1)b:tb]$ \hfill (**)
    \STATE \qquad 
    \colorbox{SpringGreen}{$w_t=w_{t-1}-\frac{\eta_s}{|\tilde{I}|}\sum_{i \in \tilde{I}} \big[\nabla f_i(w_{t-1})-(\nabla f_i(w)-\mu) \big]$}
    \STATE \qquad 
    \colorbox{BurntOrange}{ $w_t=w_{t-1}-\frac{\eta_s}{|\tilde{I}|}\sum_{i \in \tilde{I}} \big[\nabla f_i(w_{t-1})+(\nabla f_i(w)-\mu) \big]$}
    \STATE \quad \textbf{end for}
    \STATE \quad $w^{(s)}=w_{\lfloor\frac{B}{b}\rfloor}$
    \STATE \textbf{end for}
    \STATE \textbf{Output:} $w^{(T)}$
    \end{algorithmic}
\end{algorithm}

Various analyses give clues on how the outer batch size $B$ and the inner batch size $b$ may respectively influence the performance of SVRG. 
In terms of the outer batch $I$, a larger $B$ brings a more accurate estimation of the full gradient.
\cite{harikandeh2015stopwasting} argued that the full gradient estimation with an error $\epsilon$ which decreases exponentially through the training still works for SVRG without changing the theoretical convergence result.
However, the algorithm in \cite{lei2017non} with an outer batch size smaller than $|\mathcal{S}|$ additionally poses a constraint on the variance of stochastic gradients for the SVRG-like convergence results.
In this case, SVRG variants with larger $B$ are expected to converge better and enjoy better generalization.
When it comes to the inner batch $\tilde{I}$, a larger $b$ behaves similarly with a larger batch in SGD.
As mentioned in \cite{keskar2016large}, a larger $b$ may give guidance to a deeper local minimal which generalizes worse.
Therefore, a larger inner batch $\tilde{I}$ is likely to deteriorate the final generalization performance.

\renewcommand{\arraystretch}{1}
\setlength{\tabcolsep}{1em}
\begin{table*}[!htb]
\centering
\resizebox{1\textwidth}{!}{%
% \begin{tabular}{p{0.2\textwidth} p{0.2\textwidth} p{0.2\textwidth} p{0.2\textwidth} p{0.2\textwidth}}
\begin{tabular}{l|cccc}
\toprule
\diagbox[innerwidth=0.5cm]{b}{B}&50000&10000&5000&1000     \\
\midrule
10&\textbf{70.5$\pm$0.4} ($7.13h$)&69.9$\pm$0.3 ($7.18h$)&69.7$\pm$0.2 ($7.06h$)&69.0$\pm$0.3 ($6.98h$)\\
50&\textbf{70.0$\pm$0.1} ($2.21h$)&69.6$\pm$0.4 ($2.20h$)&69.1$\pm$0.3 ($2.20h$)&69.1$\pm$0.2 ($2.15h$)\\
100&\textbf{70.0$\pm$0.1} ($1.63h$)&69.7$\pm$0.1 ($1.63h$)&69.1$\pm$0.3 ($1.64h$)&68.7$\pm$0.2 ($1.63h$)\\
500&\textbf{68.6$\pm$0.2} ($1.39h$)&68.6$\pm$0.1 ($1.39h$)&68.3$\pm$0.4 ($1.39h$)&\textcolor{red}{68.8$\pm$0.1 ($1.39h$)}\\
1000&\textbf{68.0$\pm$0.2} ($1.37h$)&67.8$\pm$0.1 ($1.37h$)&67.9$\pm$0.4 ($1.37h$)&\ \ \textcolor{red}{69.1$\pm$0.08 ($1.37h$)}\\
% \midrule
\bottomrule
\end{tabular}
}
\caption{AlexNet trained on CIFAR10 for 500 epochs with average running time in parentheses.}
\label{Bbnodecay}
\vspace{-5ex}
\end{table*} 

Numerical results generally agree with the theoretical analysis mentioned above.
Table \ref{Bbnodecay} shows the generalization performance of SVRG variants with different batch size $B$, $b$, and fixed learning rate\footnote{0.001 is empirically the best learning rate for $B=50000$ and $b=10$, and $0.001\small{\times}\frac{b}{10}$ for other settings.}.
The SVRG variants with $B=50000$ calculate the exact full gradient for snapshot point update, and are obviously superior to other variants with same $b$ but smaller $B$.
Generally speaking, larger $B$ results in a better generalization performance, while larger $b$ indeed leads to a lower test accuracy.
However, such rules fail to explain the SVRG variants whose ratio of $\frac{B}{b}$ is close to one, represented by those items marked red.
In a word, smaller $B$ frees the algorithm from the full access of the dataset while larger $b$ shortens the actual running time for certain training epochs.
However, all these advantages come with a sacrifice in generalization, which is undesirable for a practical SVRG variant.

\subsection{Learning rate decay on SVRG}
Learning rate decay is another training technique commonly used in training neural networks.
The technique artificially decreases the size of update steps to guarantee the convergence to a local minimal.
Such a technique is necessary in both theoretical analysis and empirical practice of SGD, in order to control the variance of update steps introduced by the stochastic gradients.
Similarly, since the mini-batching technique for outer iterations in Batch-SVRG introduces uncertainty to the full gradient calculation, we expect that the learning rate decay can fulfill the generalization gap introduced by mini-batching techniques.
Specifically, the learning rate is divided by 5 at $40\%$, $60\%$ and $80\%$ of the total number of training epochs.
\renewcommand{\arraystretch}{1.2}
\setlength{\tabcolsep}{2em}
\begin{table*}[!htb]
\centering
\resizebox{0.9\textwidth}{!}{%
% \begin{tabular}{p{0.25\textwidth} p{0.15\textwidth} p{0.15\textwidth} p{0.15\textwidth} p{0.15\textwidth}}
\begin{tabular}{l|cccc}
\toprule
\diagbox[innerwidth=2cm]{b}{B}&50000&10000&5000&1000      \\
\midrule
100    & \textbf{71.2$\pm$0.4} & 70.9$\pm$0.2 & 71.0$\pm$0.2 & \ \ 70.9$\pm$0.05 \\
500    & 70.2$\pm$0.4 & 70.1$\pm$0.1 & \textbf{71.0$\pm$0.2} & 70.0$\pm$0.3  \\
1000   & 69.1$\pm$0.4 & \textbf{69.5$\pm$0.2} & 69.1$\pm$0.3 & \textcolor{red}{69.0$\pm$0.3}  \\
% \midrule
\bottomrule
\end{tabular}
}
\vspace{0.5ex}
\caption{AlexNet trained on CIFAR10 for 500 epochs with learning rate decay.}
\label{Bblrdecay}
\vspace{-3ex}
\end{table*} 

Numerical results verify the generalization improvement brought by learning rate decay.
In order to compare with the results before, we keep the initial learning rate same with that of the corresponding setting without learning rate decay.
Table \ref{Bblrdecay} shows that learning rate decay greatly increases the generalization performance for most of the SVRG variants.
For the bottom right item with $B=b=1000$, learning rate decay fails to bring any generalization improvement, which is mainly because of its degeneration from SVRG to SGD.
After applying learning rate decay, the difference of testing accuracy among SVRG variants with same $b$ is decreased, which indicates that the generalization improvement is greater for smaller $B$.
In this way, learning rate decay fulfills the generalization gap introduced by mini-batching techniques.
Finally, when utilizing the same inner batch size $b$ and applying learning rate decay, B-SVRG with $B<|\mathcal{S}|$ has comparable generalization performance with the corresponding SVRG variants with $B=|\mathcal{S}|$ in deep neural networks.

%% file: generalization.tex
\vspace{-0.1in}
\section{Generalization and optimization}\label{go}
Generalization and optimization are generally analyzed independently.
Analysis about optimization focuses on $\|\nabla F(w)\|^2$, since smaller $\|\nabla F(w)\|^2$ indicates a  $w$ closer to the global optimal $w^*$;
analysis about generalization supposes $w$ with flatter landscape of the objective function as a solution with better generalization performance.
However, does the optimization information tell us something about generalization?
The answer is positive, and we regard the average norm of gradients on each training sample and the norm of average gradient as metrics of the generalization ability.

\subsection{Preliminaries}
For convenience of narration, we firstly give some notions and assumptions. 
Suppose we work in the standard supervised learning setting. 
We have an unknown joint probability distribution $\mathcal{D}$ of input space $\mathcal{X}$ and target space $\mathcal{Y}$. 
We receive training data $\mathcal{S}=\{(x_1,y_1),\dots,(x_n,y_n)\}$ drawn i.i.d. from $\mathcal{D}$. 
Let $h(x,w)$ denote the output from some machine learning models with parameters $w$, which are neural networks in our context.
$f:\mathcal{Y}\rightarrow \mathcal{Y}$ is the loss function and $f(h(x,w),y)$ designate the loss of the model described by $w$ encountered on the example $(x,y)$. 
We use the notion $f_i(w)$ to denote $f(h(x_i,w),y_i)$, $f_\xi(w)$ to denote $f(h(x,w),y)$ with $\xi=(x, y)$ sampled from $\mathcal{D}$. 
The population risk and empirical risk are defined as:
\begin{equation}
    \mathcal{F}(w) = \mathbb{E}_{(x,y)\sim \mathcal{D}} [f(h(x,w),y)], F(w) =\frac{1}{n} \sum_{i=1}^{n}f_i(w).
\end{equation}
Then the generalization error of a model with parameters $w$ is the difference $\mathcal{F}(w)- F(w)$.

Throughout this paper, we denote $\|\cdot\|$ the Euclidean norm and consider that the loss function satisfies the P-L condition \cite{polyak1963gradient}. 
It is somewhat weaker than strong convexity and other popular conditions in the  literature; more extensive discussion see \cite{karimi2016linear}.
\begin{assume}\label{ass1}
$f(x)$ satisfies the P-L condition with $\mu>0$, if for any $x$
\begin{equation}
    \|\nabla f(x)\|^2 \geq 2\mu (f(x)-f(x^*)),   
\end{equation}
where $x^*$ is the global minimum of $f$.
\end{assume}
Note that a function satisfying P-L condition needs not to be convex, while a $\mu$-strongly convex function satisfies the P-L condition with $\mu$. 

\subsection{Sharpness representation under data relevant distribution}
The notion of expected sharpness as a generalization measure is recently suggested by \cite{neyshabur2017exploring} and corresponds to robustness to adversarial perturbations following certain distribution, typically a normal Gaussian distribution. That is,
\begin{equation}
\label{exp_sharp}
    S_\sigma(w) = \mathbb{E}_{\gamma \sim \mathcal{N}(0, \sigma {\bf I}_d)}[F(w+\gamma)-F(w)].
\end{equation}
% A larger $S_\sigma(w)$ indicates a steeper landscape of the objective function around $x$.
Empirical results reveal that a local minimal with a flatter surrounding landscape is likely to generalize better.
In this way, a smaller $S_\sigma(w)$ indicates a better generalization performance.

Inspired by the expected sharpness representation, we derive a general formulation of data relevant sharpness representation.
Gaussian distribution in (\ref{exp_sharp}) is designed for a general evaluation of the landscape sharpness around $w$, and has no direct relationship with the training data.
To utilize data distribution for the sharpness distribution, we have to introduce another mapping function $\varphi:\mathcal{D} \rightarrow \mathbb{R}^d$.
We substitute the Gaussian distribution with the data distribution and derive the data relevant sharpness representation $S_{\varphi}$ as follows:
\[
S_\varphi(w)=\mathbb{E}_{\xi\sim \mathcal{D}}[F(w+\varphi(\xi))-F(w)].
\]

In terms of the selection of mapping function $\varphi$, the negative as well as the positive gradients of the loss function on a single sample are natural choices, which are formulated as:
\[
    \phi_\pm(\xi) = \pm\eta\nabla f_{\xi}(w),
\]
where $\eta$ is the learning rate.
For a fixed $\xi$, $\phi_-(\xi)$ and $\phi_+(\xi)$ represent opposite directions.
The integration of $\phi_-$ and $\phi_+$ brings symmetric distribution of directions, which is reasonable in sharpness analysis.
In this way, $S_{\phi_\pm}(w)$ is empirically upper bounded by $\hat{S}_{\phi_\pm}(w)$:
\begin{equation}
\begin{aligned}
S_{\phi_\pm}(w)=&~\mathbb{E}_{\xi\sim \mathcal{D}}[F(w-\eta\nabla f_\xi(w))-F(w)]+\mathbb{E}_{\xi\sim \mathcal{D}}[F(w+\eta\nabla f_\xi(w))-F(w)]\\
\approx&~\frac{1}{n}\sum_{i=1}^{n}[\eta^2\nabla f_i(w)^T H_w \nabla f_i(w)+o(\eta^2\|\nabla f_i(w)\|^2)]\\
\lesssim&~\frac{\eta^2\lambda_{H_w}}{n} \sum_{i=1}^{n}\|\nabla f_i(w)\|^2 = \hat{S}_{\phi_\pm}(w),
\end{aligned}
\end{equation}
where $H_w$ represents the Hessian matrix of the objective function at $w$, and $\lambda_{H_w}$ represents the largest singular value of $H_w$.
$S_{\phi_\pm}$ represents the expected sharpness of the marching direction when trained on $\mathcal{S}$.
The empirical upper bound $\hat{S}_{\phi_\pm}$ is proportional to $\mathbb{E}_i\|\nabla f_i(w)\|^2=\frac{1}{n}\sum_{i=1}^n\|\nabla f_i(w)\|^2$.
Meanwhile, common optimization methods are proved to theoretically minimize $\|\nabla F(w)\|^2$.
Since $\|\nabla F(w)\|^2$ is the same order of $\frac{1}{n}\sum_{i=1}^n\|\nabla f_i(w)\|^2$, approximate solutions $w$ to the optimization problem with small $\|\nabla F(w)\|^2$ locate in a flatter landscape in terms of $S_{\phi_\pm}$.
In this way, these solutions located in flat landscape tend to generalize well.
However, $\|\nabla F(w)\|^2$ and $\frac{1}{n}\sum_{i=1}^n\|\nabla f_i(w)\|^2$ are nonidentical, and  $\frac{1}{n}\sum_{i=1}^n\|\nabla f_i(w)\|^2$ depicts the data relevant sharpness $S_{\phi_\pm}(w)$.

\subsection{Upper bound of generalization error}
In this section, we show another view of the influence of $\|\nabla F(w)\|^2$ and $\mathbb{E}_i\|\nabla f_i(w)\|^2$ in generalization. We analyze the generalization error $\mathcal{F}(w)-F(w)$ under specific conditions.
Assume $w_i^*=\arg\min_{w} f_i(w)$, $ w_*=\arg\min_{w}\mathcal{F}(w)$, and $\mathcal{F}(w)$, $f_i(w), i=1,\ldots, n$ all satisfy the
P-L condition in Assumption \ref{ass1} with $\mu$. Then we have
\begin{equation*}
\begin{aligned}
\lvert F(w)-\mathcal{F}(w)\rvert& \leq \frac{1}{2\mu}\mathbb{E}_i\|\nabla f_i(w)\|^2+\frac{1}{n}\sum_{i=1}^{n} \lvert f_i(w_i^*)-\mathcal{F}(w_*)\rvert+\frac{1}{2\mu}\| \mathbb{E}_{\xi\sim \mathcal{D}}\nabla f_{\xi}(w)\|^2.\\
\end{aligned}
\end{equation*}
If we approximate the expectation with the training data $\mathcal{S}$, then
\begin{equation}\label{eq1}
\begin{aligned}
\lvert F(w)-\mathcal{F}(w)\rvert & \lesssim  \frac{1}{2\mu} \mathbb{E}_i\|\nabla f_i(w)\|^2 +\frac{1}{2\mu}\| \nabla F(w)\|^2+\frac{1}{n}\sum_{i=1}^{n} \lvert f_i(w_i^*)-\mathcal{F}(w_*)\rvert, \\
\lvert F(w)-\mathcal{F}(w)\rvert & \lesssim \frac{1}{\mu}\mathbb{E}_i\|\nabla f_i(w)\|^2+\frac{1}{n}\sum_{i=1}^{n} \lvert f_i(w_i^*)-\mathcal{F}(w_*)\rvert, \\
\end{aligned}
\end{equation}
where ``$\lesssim $'' means roughly less-than under some conditions. Detail derivation is shown in Appendix \ref{detail_derive}. 
From Eq.~\eqref{eq1}, since $\frac{1}{n}\sum_{i=1}^{n} \lvert f_i(w_i^*)-\mathcal{F}(w_*)\rvert$ depends on the model and dataset, so smaller $\mathbb{E}_i\|\nabla f_i(w)\|^2$ and $\|\nabla F(w)\|^2$ may lead to better generalization error. 

\begin{figure}[!t]
% \centering
\begin{subfigure}[b]{0.35\textwidth}
	\centering
	\includegraphics[width=\textwidth]{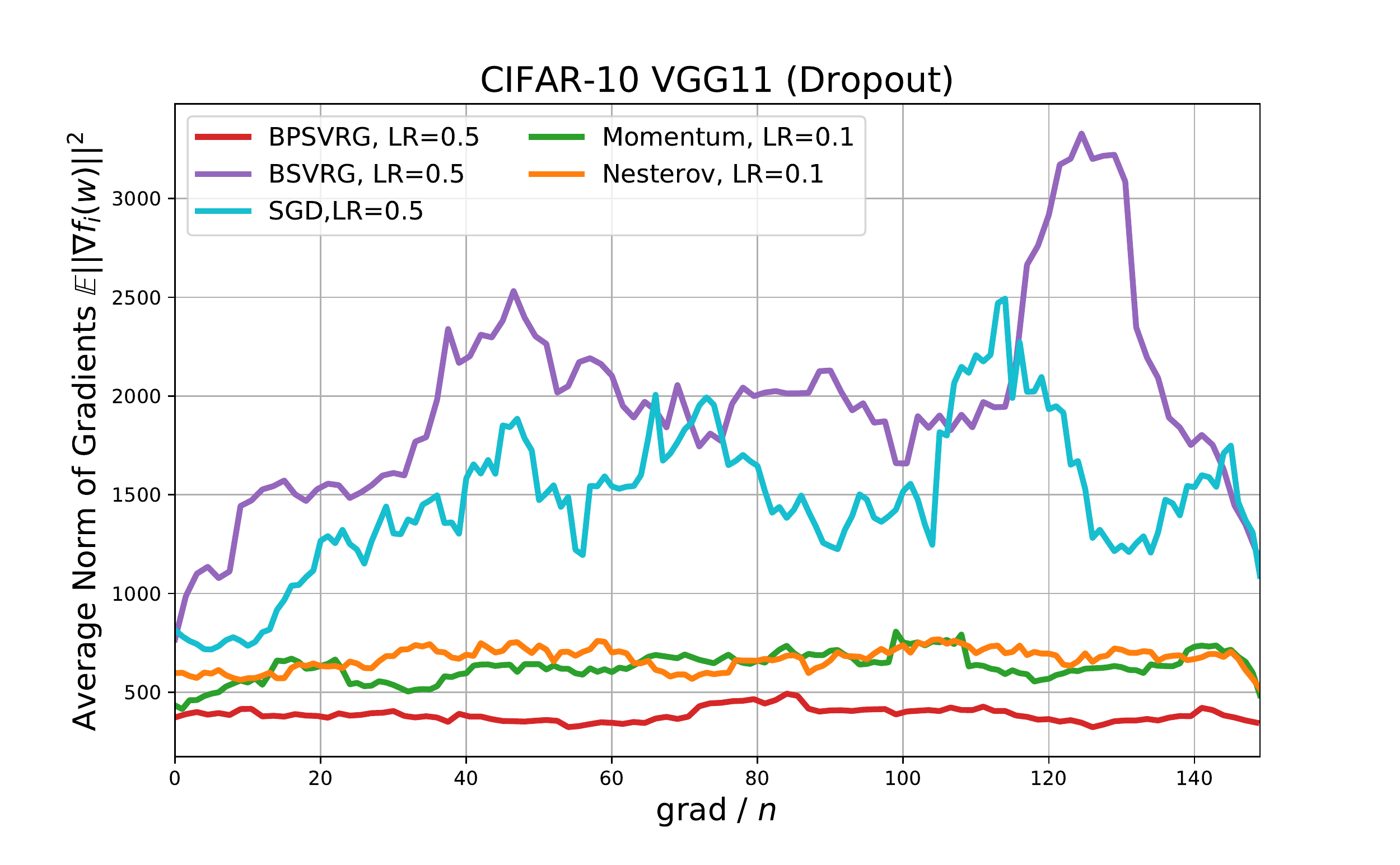}
	\vspace{-4ex}
	\caption{$\mathbb{E}_i\|\nabla f_i(w)\|^2$}
	\label{aveandfg_a}
\end{subfigure}
% \hfill
\hspace{-2cm}
 \begin{subfigure}[b]{0.35\textwidth}
	\centering
	\includegraphics[width=\textwidth]{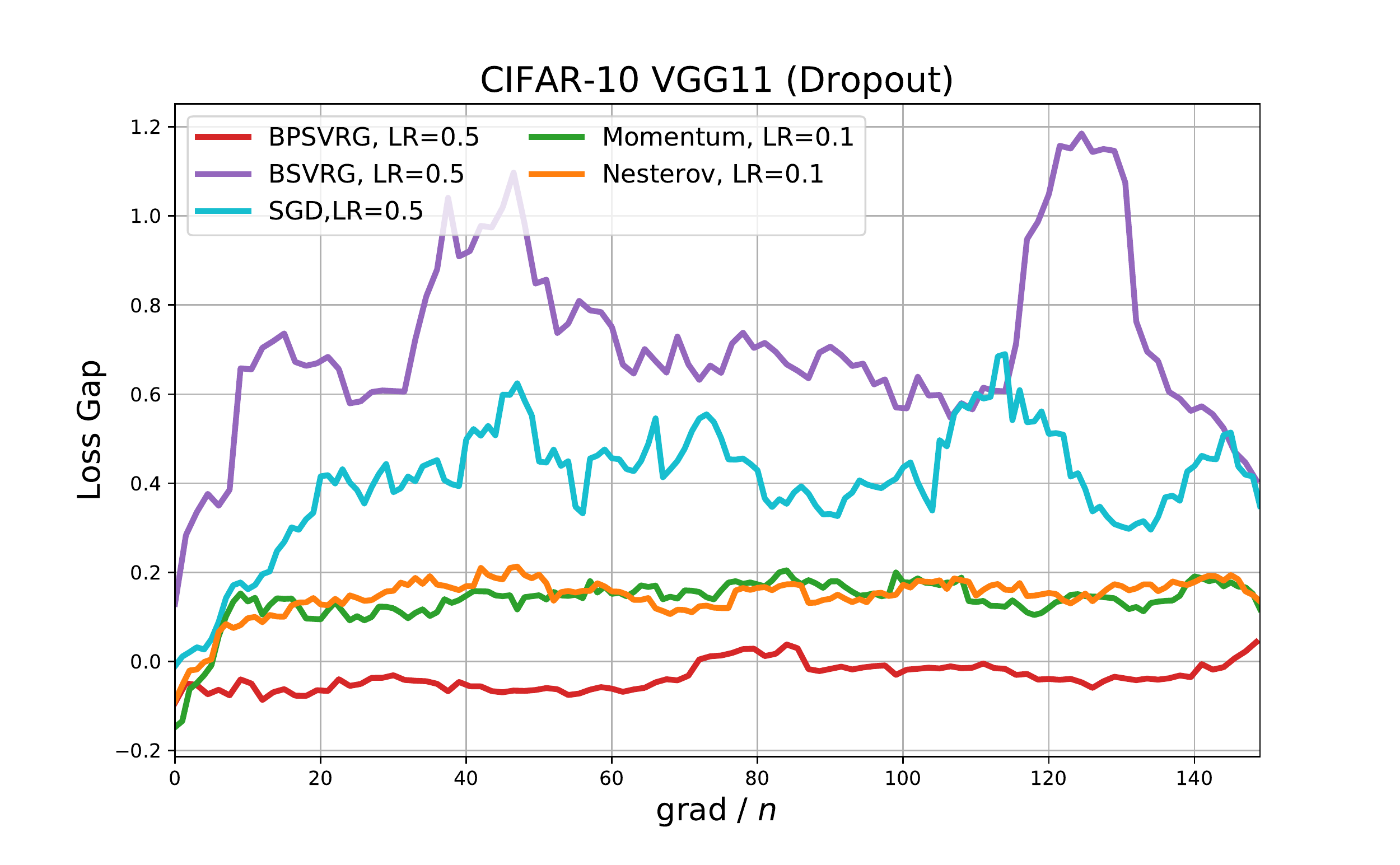}
	\vspace{-4ex}
	\caption{Loss Gap}
	\label{aveandfg_b}
\end{subfigure}
\hspace{-2cm}
% \hfill
 \begin{subfigure}[b]{0.35\textwidth}
	\centering
	\includegraphics[width=\textwidth]{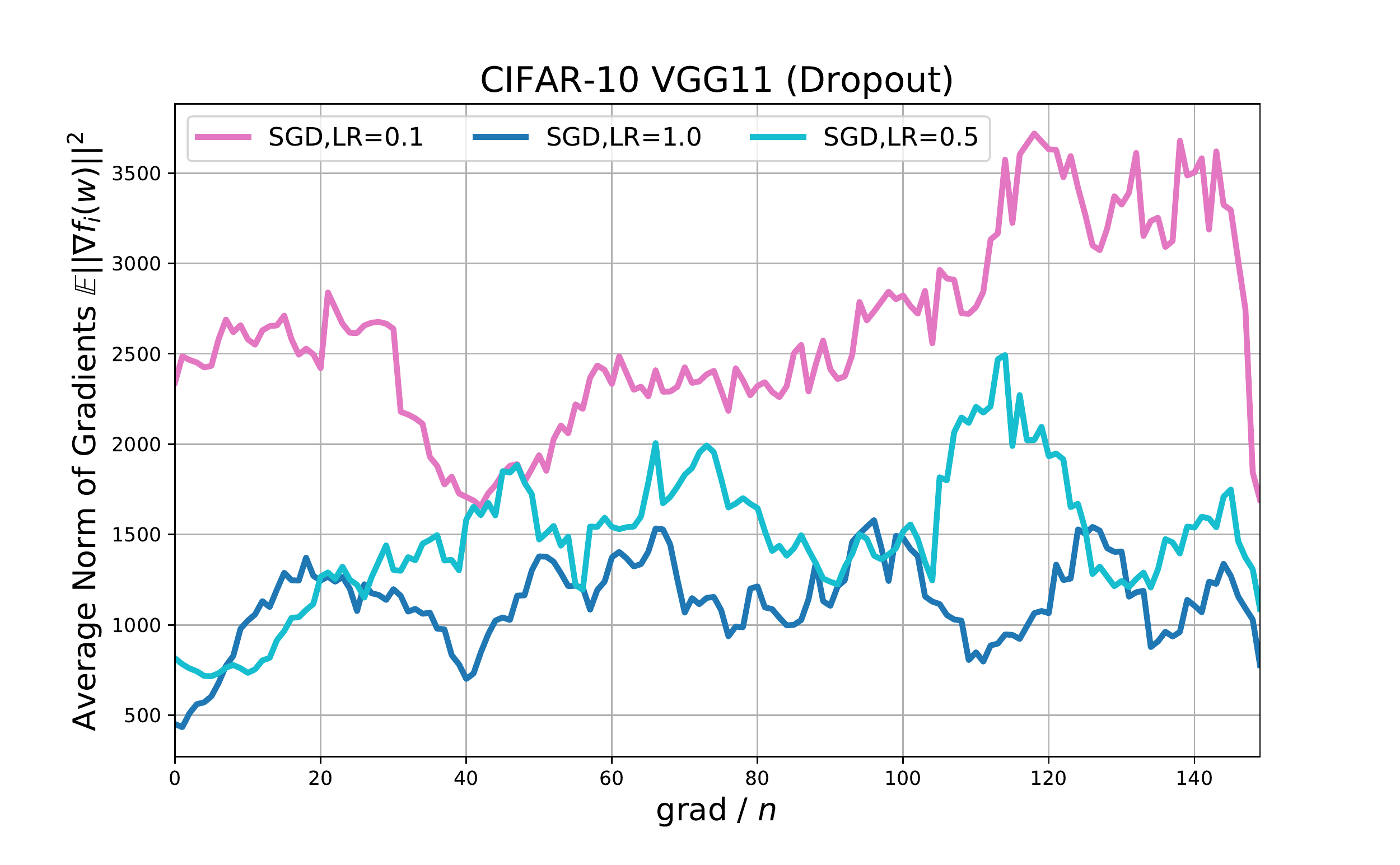}
	\vspace{-4ex}
	\caption{$\mathbb{E}_i\|\nabla f_i(w)\|^2$}
	\label{aveandfg_c}
\end{subfigure}
\begin{subfigure}[b]{0.35\textwidth}
	\centering
	\includegraphics[width=\textwidth]{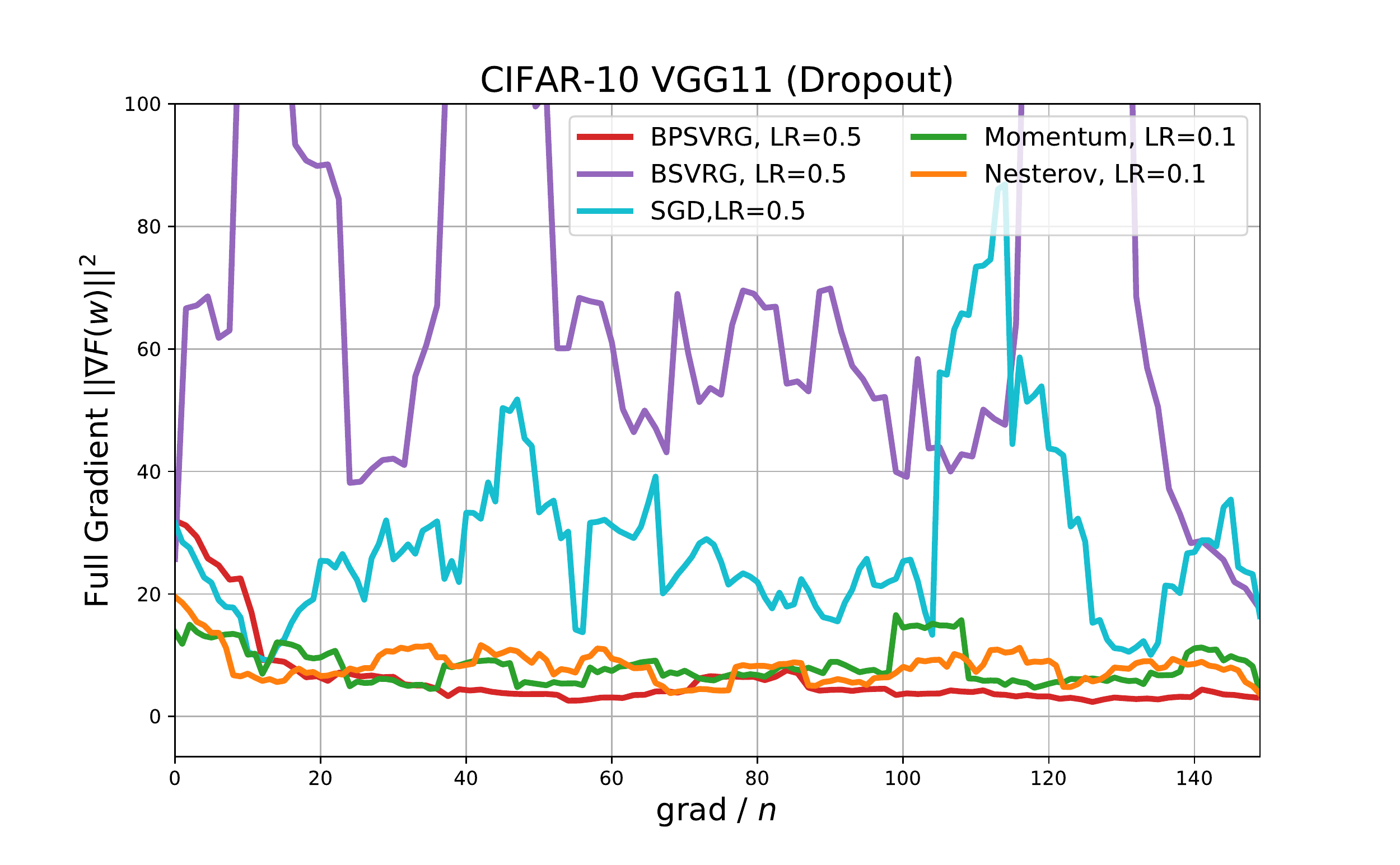}
	\vspace{-4ex}
	\caption{$\|\nabla F(w)\|^2$}
	\label{aveandfg_d}
\end{subfigure}
\hspace{-1cm}
\begin{subfigure}[b]{0.35\textwidth}
	\centering
	\includegraphics[width=\textwidth]{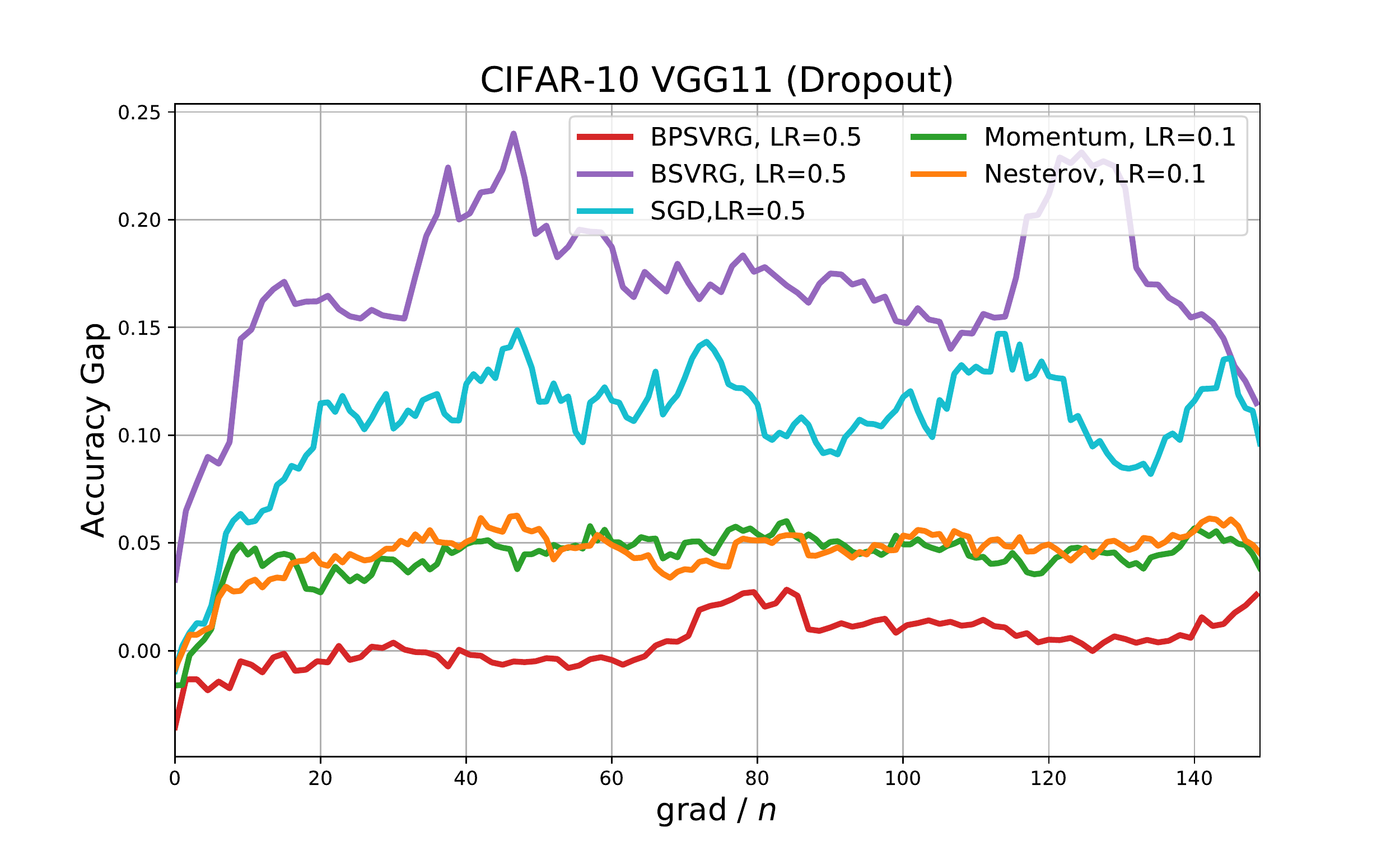}
	\vspace{-4ex}
	\caption{Accuracy Gap}
	\label{aveandfg_e}
\end{subfigure}
\hspace{-1cm}
\begin{subfigure}[b]{0.35\textwidth}
	\centering
	\includegraphics[width=\textwidth]{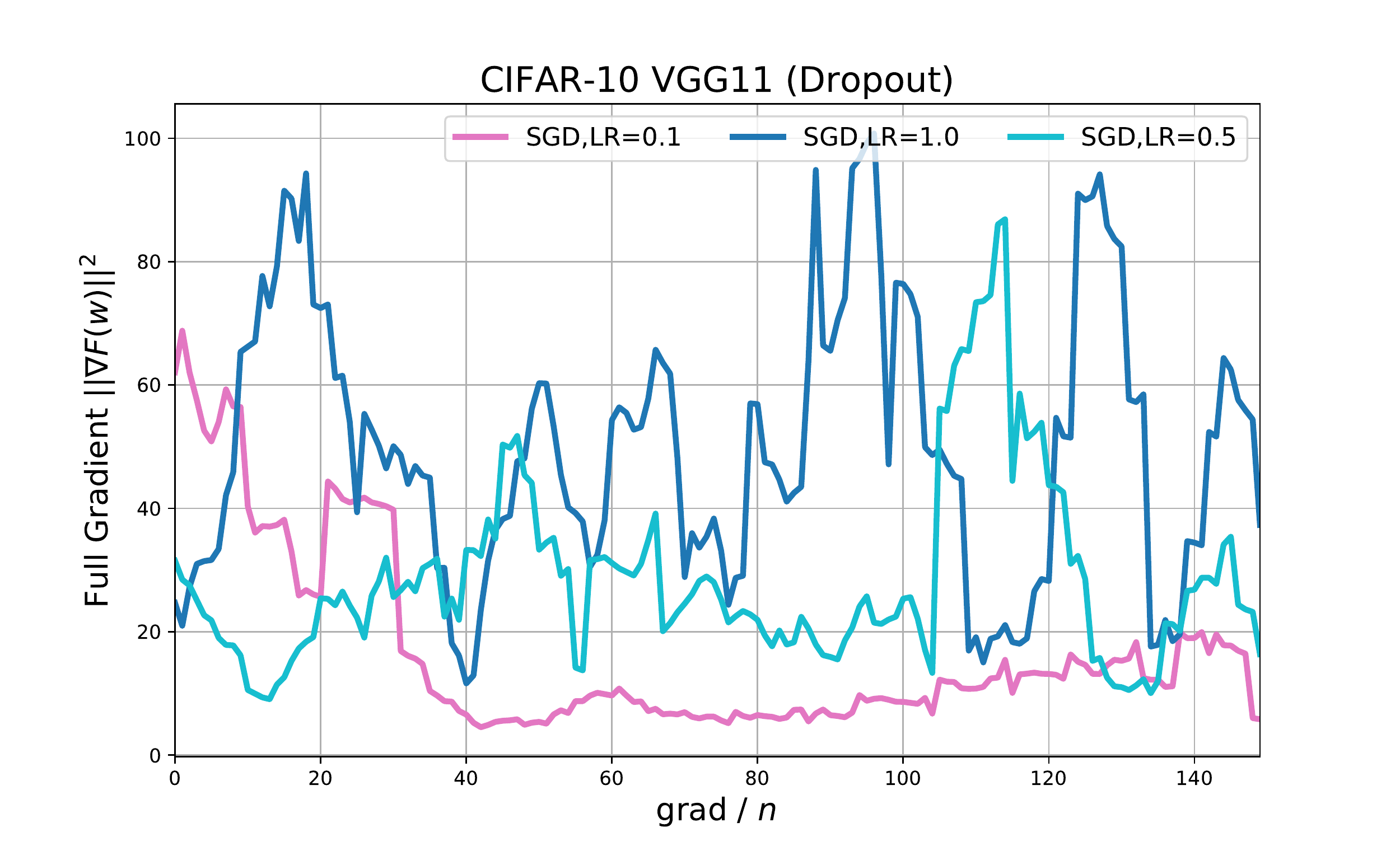}
	\vspace{-4ex}
	\caption{$\|\nabla F(w)\|^2$}
	\label{aveandfg_f}
\end{subfigure}
\vspace{-1ex}
\caption{$\mathbb{E}_i\|\nabla f_i(w)\|^2$ and $\|\nabla F(w)\|^2$ by SGD, Momentum, Nesterov momentum, B-SVRG and BP-SVRG on VGG11 with dropout on CIFAR10, without learning rate decay. Initial learning rates are shown in the legends.
Other experimental settings are same as Section \ref{num-res}. Curves are smoothed with a moving average over 5 points. 
Loss gap in (b) refers to testing loss minus training loss. 
Accuracy gap in (e) displays the difference between training and testing accuracy. 
Similar experimental figures on other networks are shown in Appendix \ref{supp_var_figure}.}
\label{aveandfg}
\vspace{-2ex}
\end{figure}

\subsection{Observation and derivation for BP-SVRG}
Based on our understanding from previous consideration, we display several empirical observations on  $\mathbb{E}_i\|\nabla f_i(w)\|^2$ and $\|\nabla F(w)\|^2$, and explain the reasons why the sign is switched in BP-SVRG. 

\textbf{Towards smaller generalization gap.}
$\mathbb{E}_i\|\nabla f_i(w)\|^2$ and $\|\nabla F(w)\|^2$ are recorded throughout the training process for different optimization methods.
Figures \ref{aveandfg}(a) and \ref{aveandfg}(d) reveal that the momentum and Nesterov enjoy smaller values of both metrics, which indicates a smaller gap for both the testing accuracy and testing loss shown in Figures \ref{aveandfg}(b) and \ref{aveandfg}(e).
In this way, $\mathbb{E}_i\|\nabla f_i(w)\|^2$ together with $\|\nabla F(w)\|^2$ indeed explain the difference between generalization gaps of different optimization methods.

\textbf{Towards flatter landscape.}
SGD with larger learning rate is shown to have better generalization, because the larger learning rate affects an SGD step similarly with smaller batch size \cite{bjorck2018understanding,hoffer2017train, smith2017bayesian,goyal2017accurate} and leads to a wider minimal.
The wider minimal is more likely to locate in a flatter landscape, and our empirical results confirm it.
As shown in Figures \ref{aveandfg}(c) and \ref{aveandfg}(f), SGD with smaller learning rate has smaller $\|\nabla F(w)\|^2$ but suffers from a larger $\mathbb{E}_i\|\nabla f_i(w)\|^2$.
In this case, SGD with larger learning rate lands in a flatter landscape in terms of the data relevant sharpness representation $\hat{S}_{\phi_\pm}$.

\textbf{Explicit variance addition in BP-SVRG.}
As shown in Figures \ref{aveandfg}(a) and \ref{aveandfg}(d), B-SVRG has relatively large values of both $\mathbb{E}_i\|\nabla f_i(w)\|^2$ and $\|\nabla F(w)\|^2$.
Motivated by the utilization of control variate $\mu_{i_t}(w^s)$ in SVRG-like methods, we switch the sign and add the control variate to stochastic gradient for explicit variance addition.
Such explicit variance addition matches the methodology of enlarging learning rate in SGD, and is empirically shown to greatly reduce both of the metrics.
Thus, BP-SVRG indeed enjoys smaller gap between the accuracy and loss, and smaller $\hat{S}_{\phi_\pm}$ implies that it is marching to flatter landscape for better generalization.

%% file: exper.tex
\section{Numerical results}\label{num-res}
\renewcommand{\arraystretch}{1.1}
\setlength{\tabcolsep}{1.1em}
\begin{table*}[!t]
\centering
\resizebox{0.98\textwidth}{!}{%
\begin{tabular}{clccc}
\toprule
Dataset & \multicolumn{1}{c}{Network}    &   BP-SVRG    &     SGD (NAG)       &      B-SVRG         \\
\midrule
CIFAR10&VGG11                &  \bf{92.98$\pm$0.14 (0.5)}  &  92.52$\pm$0.05  &  91.96$\pm$0.24    \\
&VGG11 (Dropout)             &  \bf{92.94$\pm$0.16 (0.5)}  &  92.60$\pm$0.10  &  92.00$\pm$0.23    \\
&VGG19 (Dropout)             &  \bf{94.28$\pm$0.18 (0.5)}  &  93.86$\pm$0.08  &  92.93$\pm$0.45    \\
&ResNet32                    &      93.60$\pm$0.13 (0.5)   &  93.60$\pm$0.30  &  92.71$\pm$0.09    \\
&ResNet56                    &  \bf{94.38$\pm$0.14 (0.5)}  &  94.14$\pm$0.19  &  93.24$\pm$0.09    \\
&DenseNet ($d=40, k=12$)     &      94.86$\pm$0.13 (1.0)   &  94.69$\pm$0.08  &  93.93$\pm$0.13    \\    
&DenseNet-BC ($d=100, k=12$) &  \bf{95.43$\pm$0.08 (1.0)}  &  95.23$\pm$0.12  &  94.48$\pm$0.11    \\    
\midrule
CIFAR100&VGG11 (Dropout)     &  \bf{71.22$\pm$0.25 (0.5)}  &  70.87$\pm$0.14  &  69.85$\pm$0.33    \\
&VGG19 (Dropout)             &  \bf{73.79$\pm$0.25 (0.5)}  &  73.59$\pm$0.37  &  73.25$\pm$0.10    \\
&ResNet32                    &      71.24$\pm$0.32 (1.0)   &  71.22$\pm$0.13  &  69.72$\pm$0.29    \\
&ResNet56                    &  \bf{73.12$\pm$0.34 (1.0)}  &  72.76$\pm$0.10  &  70.66$\pm$0.34    \\
&DenseNet ($d=40, k=12$)     &  \bf{75.67$\pm$0.25 (1.0)}  &  74.93$\pm$0.20  &  73.59$\pm$0.23    \\    
&DenseNet-BC ($d=100, k=12$) &  \bf{77.89$\pm$0.18 (1.0)}  &  77.51$\pm$0.28  &  75.99$\pm$0.27    \\
\midrule
SVHN&VGG11 (Dropout)         &  \bf{94.97$\pm$0.03 (0.5)}  &  94.75$\pm$0.09  &  94.39$\pm$0.06    \\
&VGG19 (Dropout)             &  \bf{95.95$\pm$0.07 (0.5)}  &  95.74$\pm$0.18  &  95.52$\pm$0.07    \\
&ResNet32                    &      95.77$\pm$0.17 (1.0)   &  95.71$\pm$0.03  &  95.47$\pm$0.10    \\
&ResNet56                    &  \bf{96.06$\pm$0.11 (1.0)}  &  95.77$\pm$0.10  &  95.39$\pm$0.24    \\
&DenseNet ($d=40, k=12$)     &      96.23$\pm$0.08 (1.0)   &  96.18$\pm$0.07  &  95.99$\pm$0.15    \\
&DenseNet-BC ($d=100, k=12$) &      96.48$\pm$0.13 (1.0)   &  96.44$\pm$0.09  &  95.77$\pm$0.05    \\
\bottomrule
\end{tabular}
}
\vspace{-0.5ex}
\caption{Comparison of testing accuracy in different methods on CIFAR and SVHN. 
All networks adopt Batch Normalization \cite{ioffe2015batch}, only VGG use dropout \cite{srivastava2014dropout} when specified, and $k, d$ are the growth rate and depth of DenseNet. 
We run 5 and 4 times on CIFAR and SVHN, and show \textit{`Best (mean$\pm$std)'} as in \cite{srivastava2015training}. 
BP-SVRG and BSVRG share same initial learning rates in the bracket.}
\label{ex_res}
\vspace{-0.5cm}
\end{table*}

In this section, we mainly display the empirical generalization performance of BP-SVRG against B-SVRG and SGD with Nesterov momentum.
For the generality of our results, these optimization methods are challenged with classification tasks of several benchmark datasets on several popular models of neural networks.
% In this paper our principal focus has been to develop an approach similar to SVRG but generates well, compare to common methods in neural network training. We empirically investigate the effectiveness of BP-SVRG on several neural network architectures, as well as several benchmark datasets. Moreover, we also show B-SVRG in some cases intuitively. 

\subsection{Experiment settings}
\textbf{Models and Datasets.}
The network architectures we use include deep convolutional networks VGG \cite{simonyan2014very}, ResNet \cite{he2016deep} and DenseNet \cite{huang2017densely}. We evaluate these models on standard deep learning datasets: CIFAR \cite{krizhevsky2009learning} and SVHN \cite{netzer2011reading}. 

\textbf{Optimization methods.}
We train baselines of all the networks using stochastic gradient descent with Nesterov momentum (NAG) \cite{sutskever2013importance} of 0.9 without dampening with the initial learning rate as 0.1. 
Specifically, the outer batch size $B$ is set to be twice of the inner batch size $b$.
Considering the update pattern of B(P)-SVRG, one additional gradient computation is required for the parameter update on one outer batch.
For the comparison fairness, we compare B(P)-SVRG of $N$ epochs with SGD (NAG) of $1.5N$ epochs to guarantee the same times of gradient computation. 

\textbf{Other Settings.}
We display the better result of BP-SVRG with initial learning rate shown in parenthesis.
In terms of the base number $N$ of training epochs, it is set to 40 on SVHN dataset for all models, while it is relatively set to 200 and 250 on CIFAR datasets for DenseNet and other models.
Detailed implementations such as learning rate decay strategy are left in Appendix \ref{sec-full-version}.

% On CIFAR datasets, except DenseNet, we train using batch size 128 for 250 epochs in B(P)-SVRG, then SGD (Nesterov) runs 375 epochs respectively for gradients computing balance.
% Both the learning rate is divided by $10$ at $40\%$, $60\%$ and $80\%$ of the total number of training epochs, while the initial learning rate of SGD is set to $0.1$.
% As for DenseNet, we use batch size $64$ for $200$ epochs in B(P)-SVRG, then SGD runs 300 epochs. 
% Both the learning rate is divided by $10$ at $50\%$ and $75\%$ of the total number of training epochs. We use a weight decay of $5\times 10^{-4}$ exclude DenseNet with $1\times10^{-4}$. 
% On SVHN dataset, we train models for 40 epoch in B(P)-SVRG, 60 epochs for SGD and use a weight decay of $10^{-4}$ for all networks. 
% Other settings are same as CIFAR. 
% On both datasets, we start with learning rate $0.5$ or $1.0$ to fine tune BP-SVRG and use the same learning rate for B-SVRG. Detail implementations are displayed in Appendix \ref{sec-full-version}. 

\subsection{Generalization analysis}
\textbf{Testing accuracy.}
The main results are shown in Table \ref{ex_res}. 
For almost all the network structures, B-SVRG is inferior to SGD (NAG), and BP-SVRG enjoys comparable performance with SGD(NAG).
To highlight general trends, we mark all results in which BP-SVRG outperforms SGD (NAG) obviously in boldface.
Moreover, in terms of dataset traversal time, BP-SVRG is apparently superior to SGD (NAG).
Specifically, in our settings, SGD actually traverses the dataset $50\%$ more than BP-SVRG.
In this way, BP-SVRG is preferable for scenarios where dataset traversal is time consuming.

\begin{figure}[!t]
    \centering
	\includegraphics[width=0.35\linewidth]{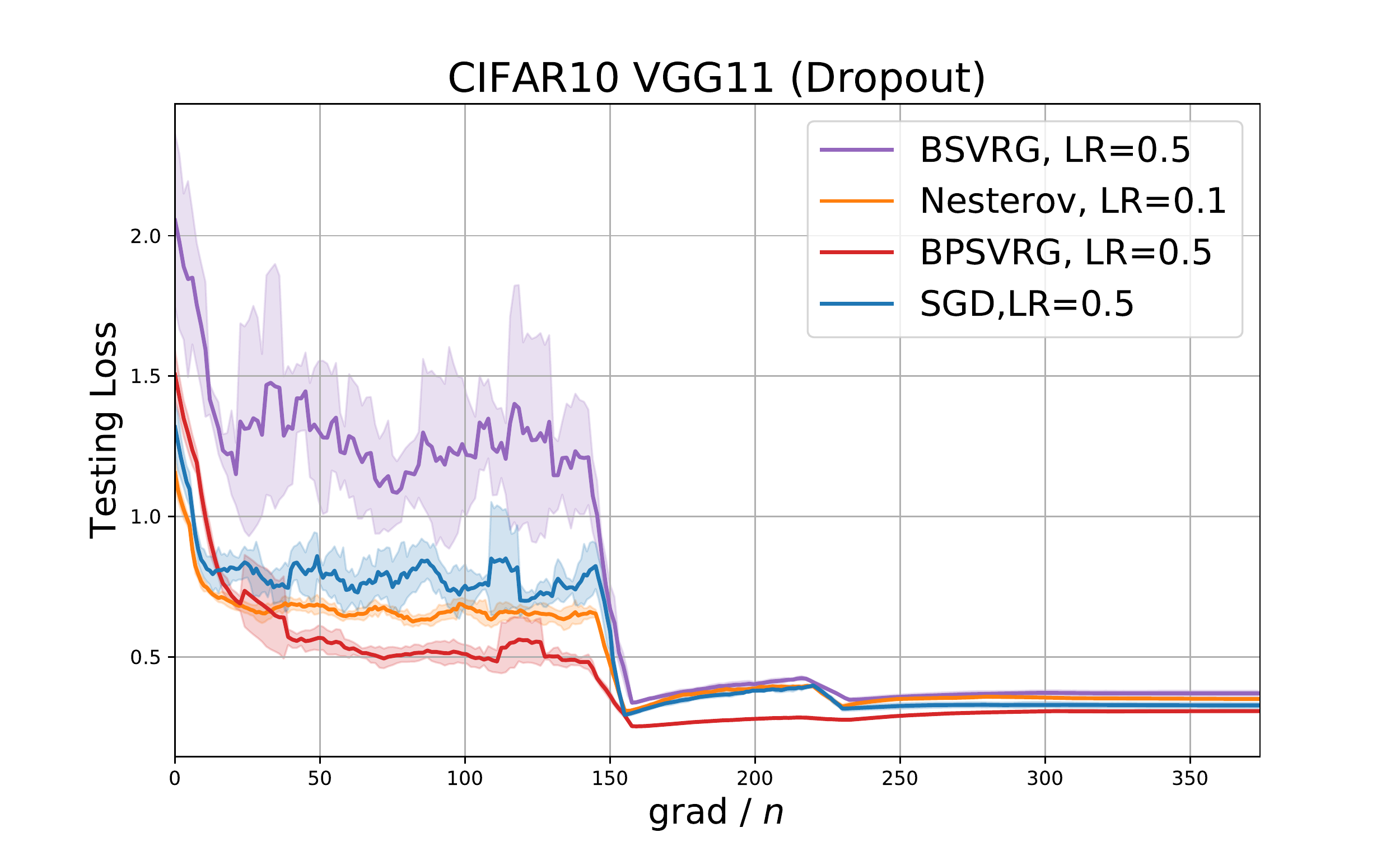}
	\hspace{-0.6cm}
    \includegraphics[width=0.35\linewidth]{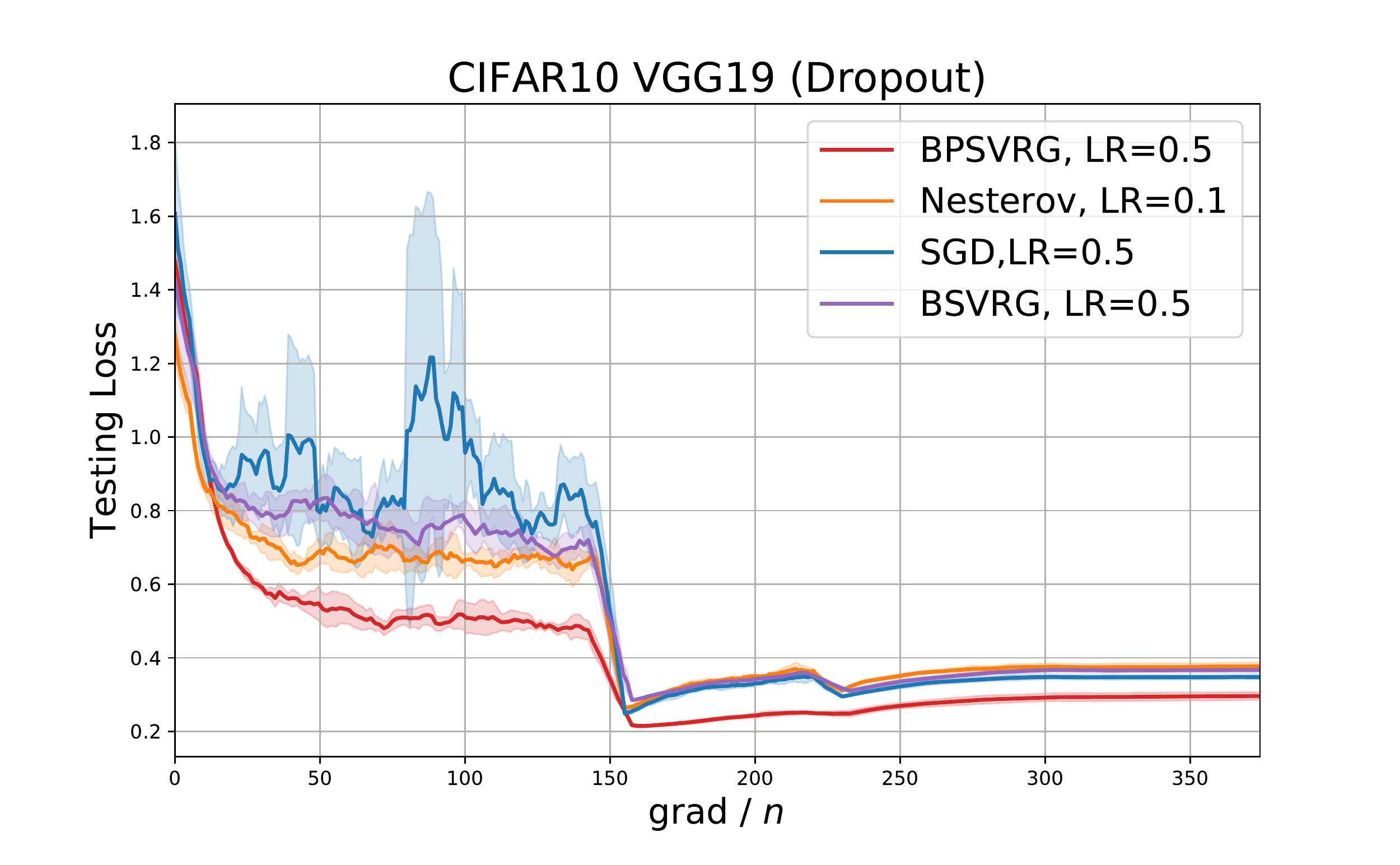}
    \hspace{-0.6cm}
    \includegraphics[width=0.35\linewidth]{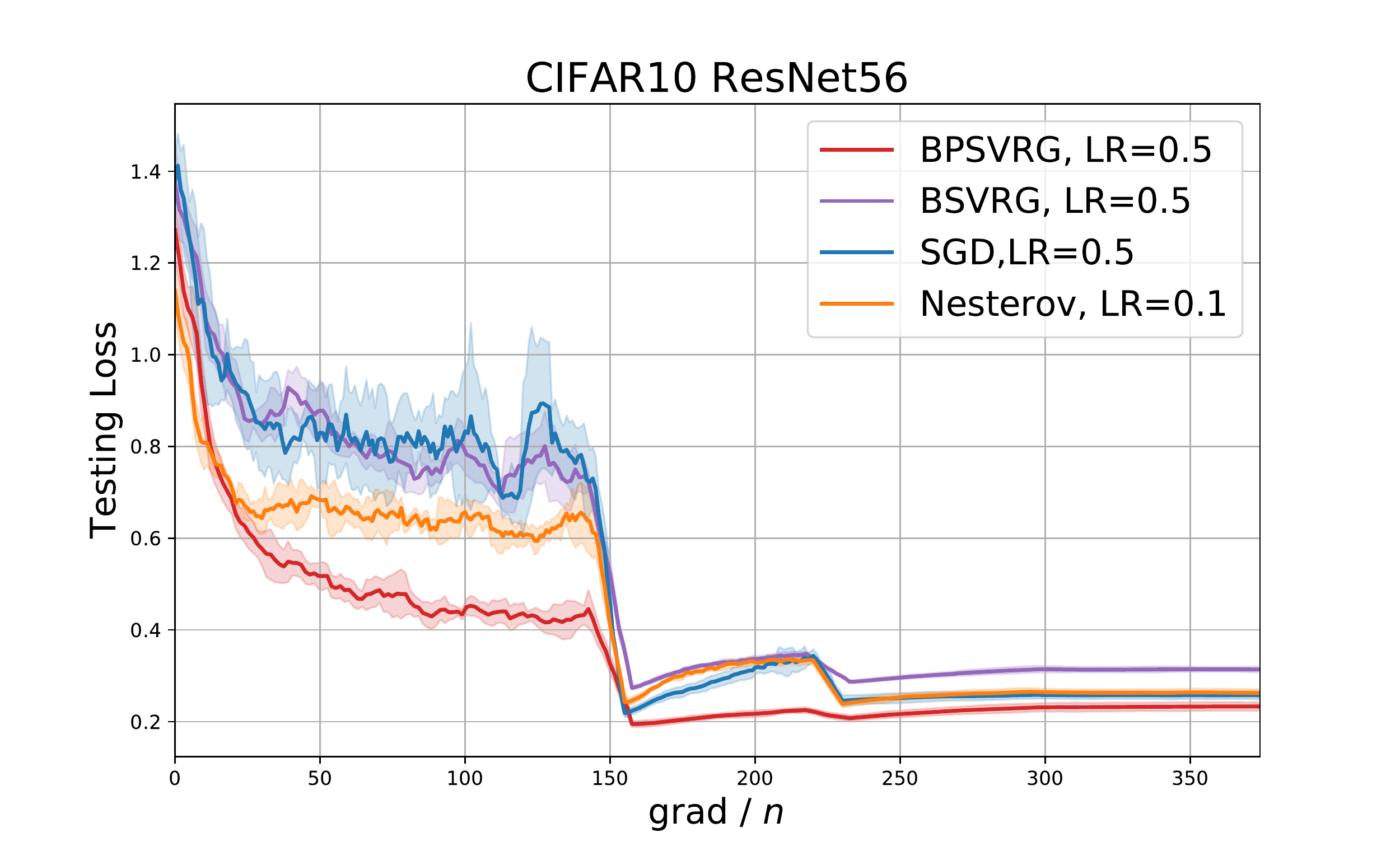}
    \vspace{1pt}
	\caption{Testing loss with std on VGG and ResNet in CIFAR-10. Here testing loss is the softmax-cross-entropy of network output with true label, without regularization term. Curves are smoothed with a moving average over 5 points.}
	\label{testloss1}
	\vspace{-0.2in}
\end{figure}

\textbf{Testing loss.}
Testing loss is a more direct metric for the generalization of optimization methods.
As is shown in Figure \ref{testloss1}, BP-SVRG generally enjoys the smallest test loss throughout the training process.
The advantages of BP-SVRG in dataset traversal time mentioned above is clear in Figure \ref{testloss1}.
Specifically, before any learning rate decay, BP-SVRG obtains a relatively small testing loss which indicates a better generalization.
Empirical results reveal that BP-SVRG actually brings out the acceleration in generalization.
% In our opinion, such acceleration in generalization may be more important than the acceleration in optimization.
Similar phenomenon on other network structures and datasets are shown in Appendix \ref{compare_figure}. 

%% file: conclusion.tex
\section{Conclusion}
In this work we have empirically investigated widely used training techniques on SVRG-type methods: mini-batching techniques increase the scalability in terms of training data size and network structure, while learning rate decay recovers the generalization sacrifice caused by mini-batching techniques. 
Furthermore, in the scope of generalization and optimization, we have shown that the norm of average gradient and the average norm of gradients measure how the model generalizes with some insight derivation.
It is empirically observed that optimization methods with smaller values of these metrics enjoy better generalization performance.
Through switching the sign in B-SVRG, we perform explicit variance addition to the stochastic gradient and derive BP-SVRG, which has smaller values in our metrics and empirically enjoy better generalization performance.

As an optimization method for deep neural networks, BP-SVRG reuses batches of data and is empirically shown to obtain smaller test loss with fewer times of traversing the dataset.
In this way, optimization methods with the introduction of snapshot points may have potentials to accelerate the generalization through data reutilization, which is left to future work.
Moreover, the average norm of gradients on each sample seems to offer a more vivid depiction of how well a deep neural network generalizes.
The interaction between the norm of average gradient and the average norm of gradients remains unclear, which may help to further understand the generalization of deep neural networks.

% We believe that the empirical performance of BP-SVRG may lead to introspection on whether small variance of stochastic gradient is desirable in deep neural networks.
% Additionally, the average norm of gradients on each sample seems to offer a more vivid depiction of how well a deep neural network generalizes.
% Although the norm of average gradient and the average norm of gradients bound each other, yet the actual correlation between these quantities remains to be an open problem.

%% file: appendix.tex
\newpage
\appendix
\section{Experiments details}\label{sec-full-version}
\subsection{Networks and devices}
Experiments are run using PyTorch \cite{paszke2017automatic} on the machine with 128 GB RAM, and NVIDIA TESLA P100 GPU.
We use VGG from PyTorch modules and cut down the neurons in full-connected layers all to $512$ instead of $4096$. 
We adopt DenseNet from \url{https://github.com/andreasveit/densenet-pytorch}, which is recommended by the author of DenseNet, and ResNet from \url{https://github.com/akamaster/pytorch_resnet_cifar10}. 
% Experiments are run by ourselves and some may not be comparable to benchmarks in the original paper.

\subsection{Other experimental settings}
In terms of the batch size, the inner batch size $b$ of B(P)-SVRG is set to be same as the batch size in SGD (NAG).
Specifically, $b$ is respectively set to 128 except for DenseNet in CIFAR datasets with 64.
The initial learning raet of SGD (NAG) is set to $0.1$ on all of the three datasets.
For all of the optimization methods, the learning rate is divided by $10$ at $40\%$, $60\%$ and $80\%$ of the total number of training epochs on CIFAR datasets excluding Densenet, while the learning rate is divided by $10$ at $50\%$ and $75\%$ of the total number of training epochs on SVHN dataset and DenseNet on CIFAR.
In terms of the $\ell_2$-regularization added to the loss on the networks, the coefficient is relatively set to $10^{-4}$ for  DenseNet on CIFAR and all networks on SVHN dataset, else $5\small{\times}10^{-4}$.

% On CIFAR datasets, except DenseNet, we train using batch size 128 for 250 epochs in B(P)-SVRG, then SGD (Nesterov) runs 375 epochs respectively for gradients computing balance.
% Both the learning rate is divided by $10$ at $40\%$, $60\%$ and $80\%$ of the total number of training epochs, while the initial learning rate of SGD is set to $0.1$.
% As for DenseNet, we use batch size $64$ for $200$ epochs in B(P)-SVRG, then SGD runs 300 epochs. 
% Both the learning rate is divided by $10$ at $50\%$ and $75\%$ of the total number of training epochs. We use a weight decay of $5\times 10^{-4}$ exclude DenseNet with $1\times10^{-4}$. 
% On SVHN dataset, we train models for 40 epoch in B(P)-SVRG, 60 epochs for SGD and use a weight decay of $10^{-4}$ for all networks. 
% Other settings are same as CIFAR. 
% On both datasets, we start with learning rate $0.5$ or $1.0$ to fine tune BP-SVRG and use the same learning rate for B-SVRG. Detail implementations are displayed in Appendix \ref{sec-full-version}. 

\subsection{Dataset preprocessing}
The two CIFAR datasets, CIFAR-10 and CIFAR-100 \cite{krizhevsky2009learning} consist of colored natural images with $32\times32$ pixels with 10 and 100 classes respectively.
We adopt a standard data augmentation scheme (mirroring/cropping) that is widely used on these two datasets \cite{he2016deep, srivastava2015training, huang2017densely, huang2016deep}. 
In addition, we normalize the data using the channel means and standard deviations for preprocessing.

The Street View House Numbers (SVHN) dataset \cite{netzer2011reading} contains $32\times32$ colored digit images, 73,257 images in the training set, 26,032 images in the testing set, and 531,131 images for additional training.  
We just use the training set of 73257 images without any data augmentation, and divide the pixel values by 255 so that they are in the [0, 1] range followed by \cite{zagoruyko2016wide}.

% \subsection{Hyperparameters Details}\label{hyper}
% On CIFAR datasets, except DenseNet, we train using batch size 128 for 250 epochs in B(P)-SVRG, then SGD runs 375 epochs respectively for gradients computing balance. Both the learning rate is divided by $10$ at $40\%$, $60\%$ and $80\%$ of the total number of training epochs, while the initial learning rate of SGD is set to $0.1$.
% As for DenseNet, we use batchsize $64$ for $200$ epochs in B(P)-SVRG, then SGD plays 300 epochs. Both the learning rate is divided by $10$ at $50\%$ and $75\%$ of the total number of training epochs, while the initial learning rate of SGD is set to $0.1$. We use a weight decay of $5\times 10^{-4}$ except DenseNet and $1\times10^{-4}$ for DenseNet. 
% On SVHN dataset, we train models for 40 epoch in B(P)-SVRG, 60 epochs for SGD and other settings are same as CIFAR. We use a weight decay of $10^{-4}$ for all networks. On both dataset, we start with learning rate $0.5$ or $1.0$ to fine tune BP-SVRG and use the same learning rate for B-SVRG. 

\subsection{Batch Normalization setting}\label{bn}
For SVRG-type methods, when Batch Normalization (BN) is applied in training, it is necessary to store mean and variance for test use.
The standard approach is to keep track of an exponential moving average of the mean and variances computed at each training step. 
During training time on ResNet, we discover poor results and divergence in B(P)-SVRG when running snapshot without mean and variance modification. 
Since we compare B(P)-SVRG of $N$ epochs with SGD of $1.5N$ epochs, we empirically update the mean and variances information in the gradient computation of snapshot points to obtain comparable distribution statistics for different optimization methods.
Thus, we use training mode for BN for every gradient computation, which stablizes the training of SVRG-type methods with large learning rate. 
To guarantee that merely additional BNs do not improve the generalization performance, we also make controlled trials of SGD. 
Speicifically, we run Modified-SGD (see Algorithm \ref{mo-sgd}) which is similar to B(P)-SVRG, instead of cancelling out the control variate in the parameter update.
We compare the generalizatino performance of Modified-SGD with that of SGD. 
Other settings are same as Section \ref{num-res}.

\begin{algorithm}
    \begin{algorithmic}
    \caption{Modified-SGD}
    \label{mo-sgd}
    \STATE $\textbf{Input}$ Training dataset $\mathcal{S}$, learning rate $\{\eta_s\}$, outer batch size $B$, and inner batch size $b$ ($B\geq b$), number of epochs $T$.
    \STATE $\textbf{Initialize}$ $w^{(0)}$
    \STATE \textbf{for} $s=1,~2,~\ldots,~T$ \textbf{do}
    \STATE \quad $w_0 = w^{(s-1)}$
    \STATE \quad $I$ randomly selected from $\mathcal{S}$ with $|I|=B$ \hfill
    \STATE \quad \textbf{for} $t=1,~2,~\ldots,~\lfloor\frac{B}{b}\rfloor$ \textbf{do}
    \STATE \quad \quad $\tilde{I}=I[(t-1)b:tb]$ \hfill
    \STATE \quad \quad Run $\frac{1}{|\tilde{I}|}\sum_{i\in \tilde{I}}^{}f_i(w)$ to keep track of mean and variances in BN
    \STATE \quad \textbf{end for}
    \STATE \quad \textbf{for} $t=1,~2,~\ldots,~\lfloor\frac{B}{b}\rfloor$ \textbf{do}
    \STATE \quad \quad $\tilde{I}=I[(t-1)b:tb]$ \hfill
    \STATE \qquad $w_t=w_{t-1}-\frac{\eta_s}{|\tilde{I}|}\sum_{i \in \tilde{I}}\nabla f_i(w_{t-1})$
    \STATE \quad \textbf{end for}
    \STATE \quad $w^{(s)} = w_{\lfloor\frac{B}{b}\rfloor}$
    \STATE \textbf{end for}
    \STATE \textbf{Output:} $w^{(T)}$
    \end{algorithmic}
\end{algorithm}

\begin{table*}[!ht]
\centering
\resizebox{0.65\textwidth}{!}{
\begin{tabular}{p{0.25\textwidth}cc}
\toprule
{}                        &   Modified-SGD  &     SGD(NAG)    \\
\midrule
VGG11 (Dropout)           &  92.45$\pm$0.18 & 92.60$\pm$0.10  \\
Resnet56                  &  93.93$\pm$0.14 & 94.14$\pm$0.19  \\
% \midrule
\bottomrule
\end{tabular}
}
\vspace{0.5ex}
\caption{Compare accuracy of different BN alternatives on CIFAR10.}
\label{modi-sgd}
\vspace{-1ex}
\end{table*} 
As is shown in Table \ref{modi-sgd}, repeated running statistic in BN makes the generalization performance even worse, regardless of dropout; hence we confirm the effectiveness of our B(P)-SVRG methods in final experimental analysis.

\section{Detail derivation}\label{detail_derive}
This appendix shows detail derivation in Section \ref{go}.

\subsection{Upper Bound of Generalization Error}
Assume $w_i^*=\arg\min_{w} f_i(w)$, $ w_*=\arg\min_{w}\mathcal{F}(w)$, and $\mathcal{F}(w)$, $f_i(w), i=1,\dots,n$ all satisfy P-L condition in Assumption \ref{ass1} with $\mu$. 
Then we have
\begin{equation*}
\begin{aligned}
\lvert F(w)- \mathcal{F}(w)\rvert&=\bigg\lvert \frac{1}{n}\sum_{i=1}^{n}f_i(w)-\mathcal{F}(w) \bigg\rvert \leq \frac{1}{n}\sum_{i=1}^{n}\lvert f_i(w)-\mathcal{F}(w)\rvert\\
&=\frac{1}{n}\sum_{i=1}^{n}\lvert f_i(w)-f_i(w_i^*)+f_i(w_i^*)-\mathcal{F}(w_*)+\mathcal{F}(w_*)-\mathcal{F}(w)\rvert\\
&\leq \frac{1}{n}\sum_{i=1}^{n}\lvert f_i(w)-f_i(w_i^*)\rvert+\frac{1}{n}\sum_{i=1}^{n}\lvert f_i(w_i^*)-\mathcal{F}(w_*)\rvert+\lvert \mathcal{F}(w_*)-\mathcal{F}(w)\rvert\\
&\leq \frac{1}{n}\sum_{i=1}^{n} \frac{1}{2\mu} \|\nabla f_i(w)\|^2 + \frac{1}{n}\sum_{i=1}^{n} \lvert f_i(w_i^*)-\mathcal{F}(w_*)\rvert+\frac{1}{2\mu}\| \nabla \mathcal{F}(w)\|^2\\
&= \frac{1}{2\mu}\mathbb{E}_i\|\nabla f_i(w)\|^2+\frac{1}{n}\sum_{i=1}^{n} \lvert f_i(w_i^*)-\mathcal{F}(w_*)\rvert+\frac{1}{2\mu}\|\nabla \mathbb{E}_{\xi\sim \mathcal{D}}f_{\xi}(w)\|^2\\
&\approx \frac{1}{2\mu}\mathbb{E}_i\|\nabla f_i(w)\|^2+\frac{1}{n}\sum_{i=1}^{n} \lvert f_i(w_i^*)-\mathcal{F}(w_*)\rvert+\frac{1}{2\mu}\| \mathbb{E}_{\xi\sim \mathcal{D}}\nabla f_{\xi}(w)\|^2.\\
\end{aligned}
\end{equation*}
The second inequality is by P-L condition of $f_i(w)$ and $\mathcal{F}(w)$, while the approximate equality is by exchanging the expectation and the derivative in common cases.\\
If we approximate the expectation by sample $\mathcal{S}$, then
\begin{equation*}
\mathbb{E}_{\xi\sim \mathcal{D}}\nabla f_{\xi}(w) \approx \frac{1}{n}\sum_{i=1}^{n}\nabla f_i(w)=\nabla F(w).
\end{equation*}
Hence,
\begin{equation}\label{aeq1}
\lvert F(w)-\mathcal{F}(w)\rvert  \lesssim  \frac{1}{2\mu} \mathbb{E}_i\|\nabla f_i(w)\|^2 +\frac{1}{2\mu}\|\nabla F(w)\|^2+\frac{1}{n}\sum_{i=1}^{n} \lvert f_i(w_i^*)-\mathcal{F}(w_*)\rvert
\end{equation}
where $\lesssim$ means approximately less-than under some conditions. \\
Moreover, since $\|\cdot\|^2$ is convex, and use training data $\mathcal{S}$ to approximate the expectation, then  
\begin{equation}
\begin{aligned}
\| \mathbb{E}_{\xi\sim \mathcal{D}}\nabla f_{\xi}(w)\|^2 &\leq \mathbb{E}_{\xi\sim \mathcal{D}} \|\nabla f_{\xi}(w)\|^2\approx \mathbb{E}_i\|\nabla f_i(w)\|^2.
\end{aligned}
\end{equation}
Therefore,
\begin{equation}\label{aeq2}
\begin{aligned}
\lvert F(w)-\mathcal{F}(w)\rvert &  \lesssim \frac{1}{\mu}\mathbb{E}_i\|\nabla f_i(w)\|^2+\frac{1}{n}\sum_{i=1}^{n} \lvert f_i(w_i^*)-\mathcal{F}(w_*)\rvert\\
\end{aligned}
\end{equation}
Although deep neural networks may not meet the assumptions, we believe that such heuristic results are not trivial and offer us opportunities to further understand the generalization of deep neural networks.
% However, the assumptions are unsuitable for neural networks. We mention this for heuristic consideration.

\section{Testing Loss Comparison} \label{compare_figure}
\begin{figure}[!t]
	\centering
	\includegraphics[width=0.35\linewidth]{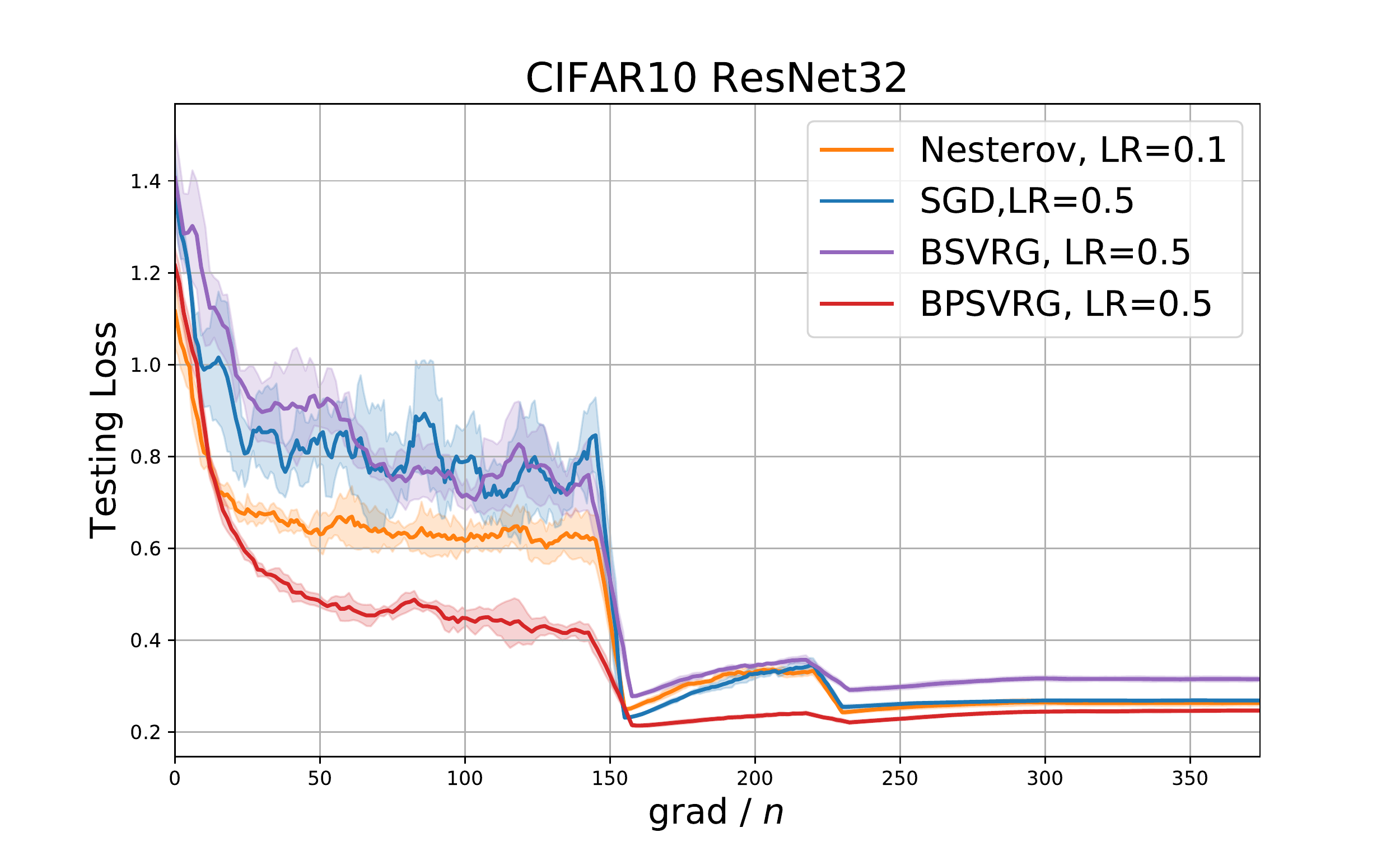}
	\hspace{-0.6cm}
	\includegraphics[width=0.35\linewidth]{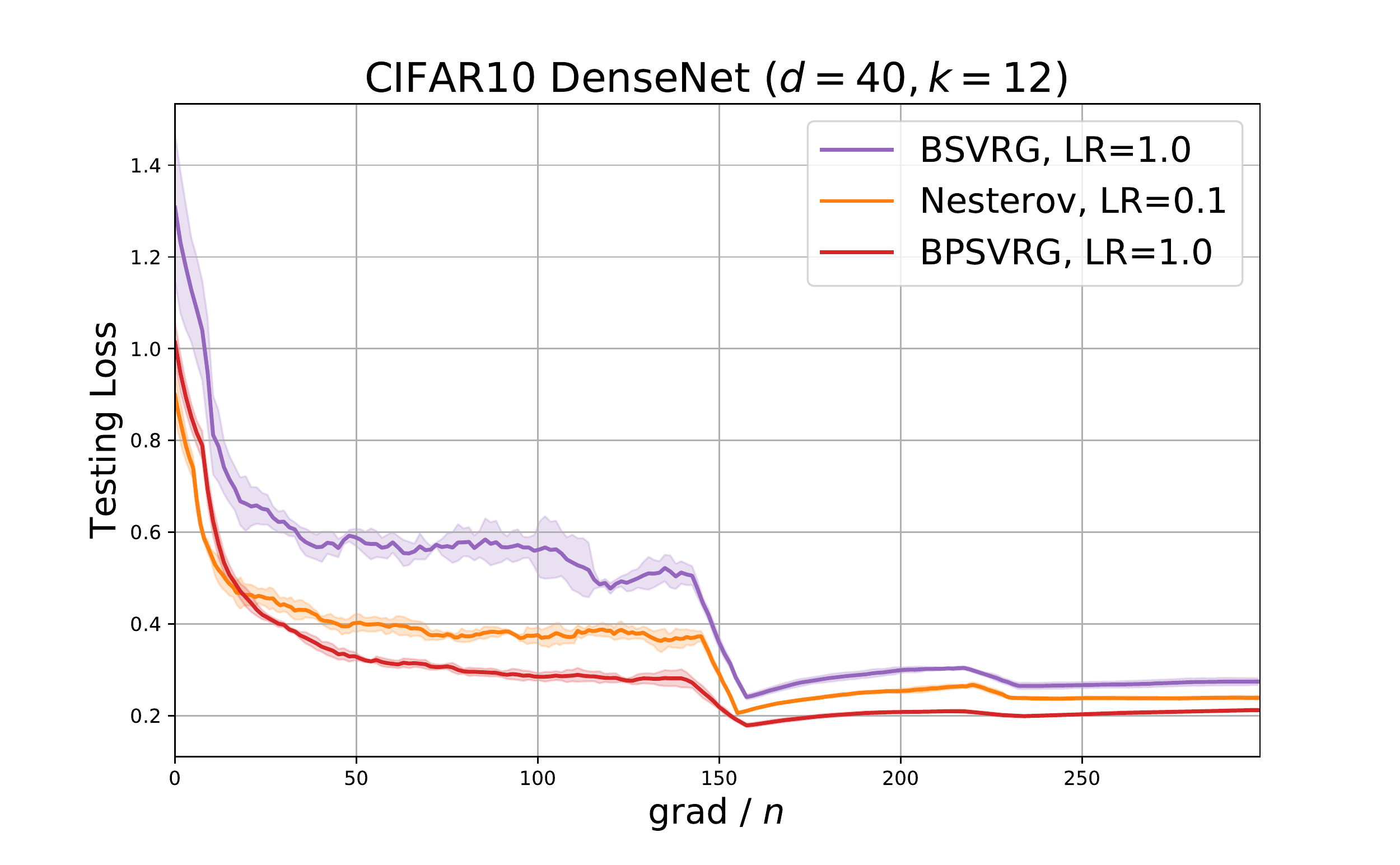}
	\hspace{-0.6cm}
	\includegraphics[width=0.35\linewidth]{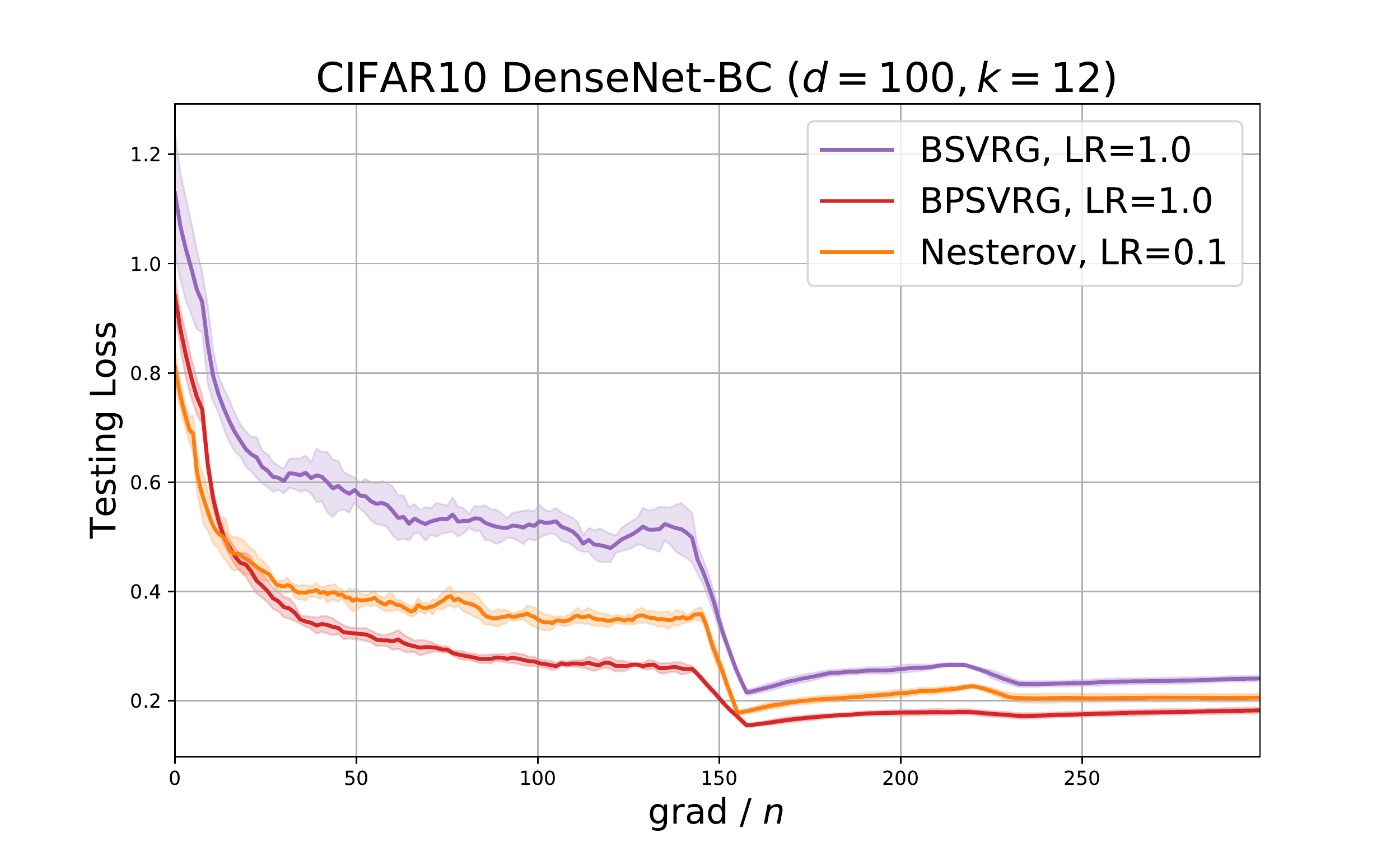}
	\includegraphics[width=0.35\linewidth]{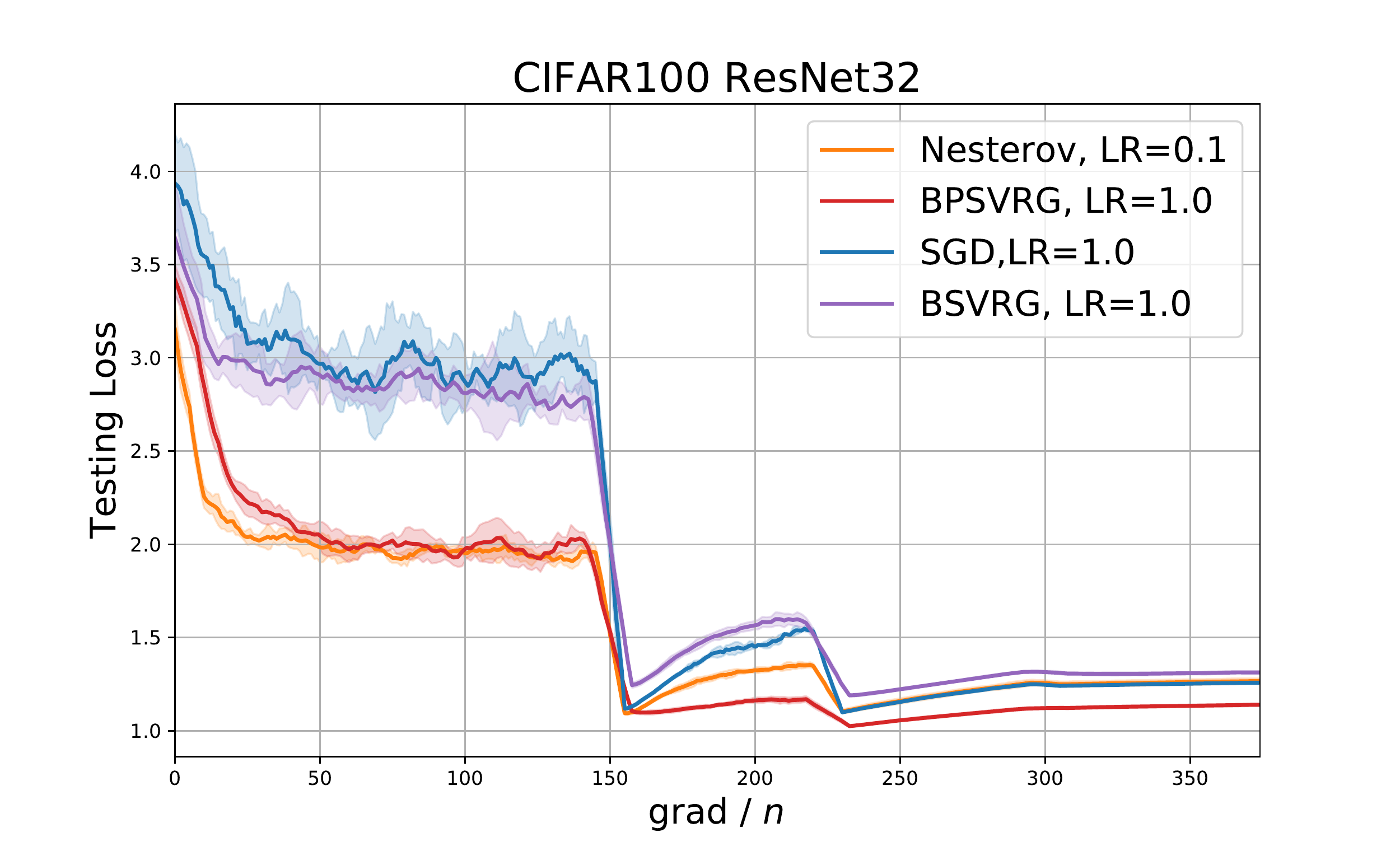}
	\hspace{-0.6cm}
	\includegraphics[width=0.35\linewidth]{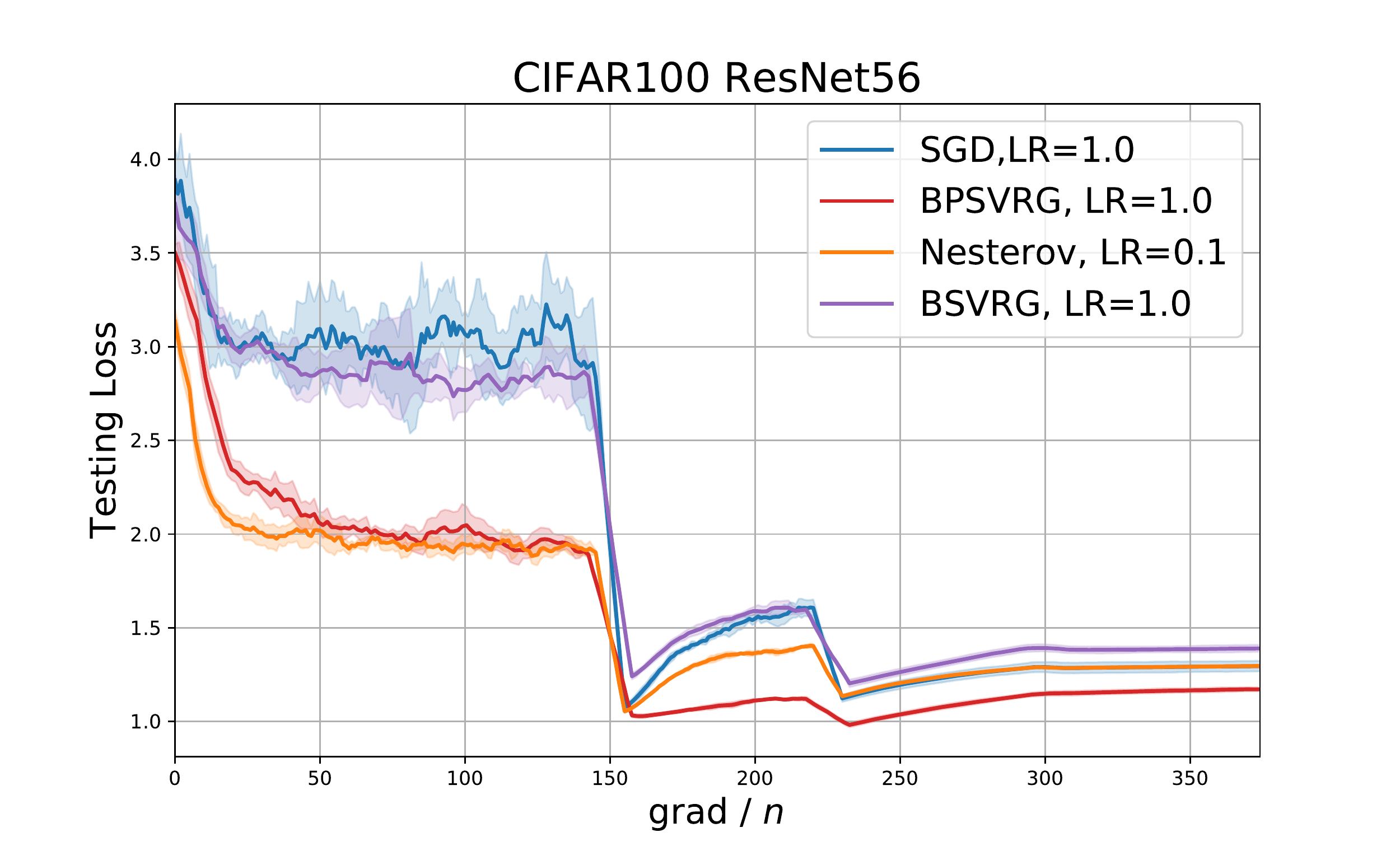}
	\hspace{-0.6cm}
	\includegraphics[width=0.35\linewidth]{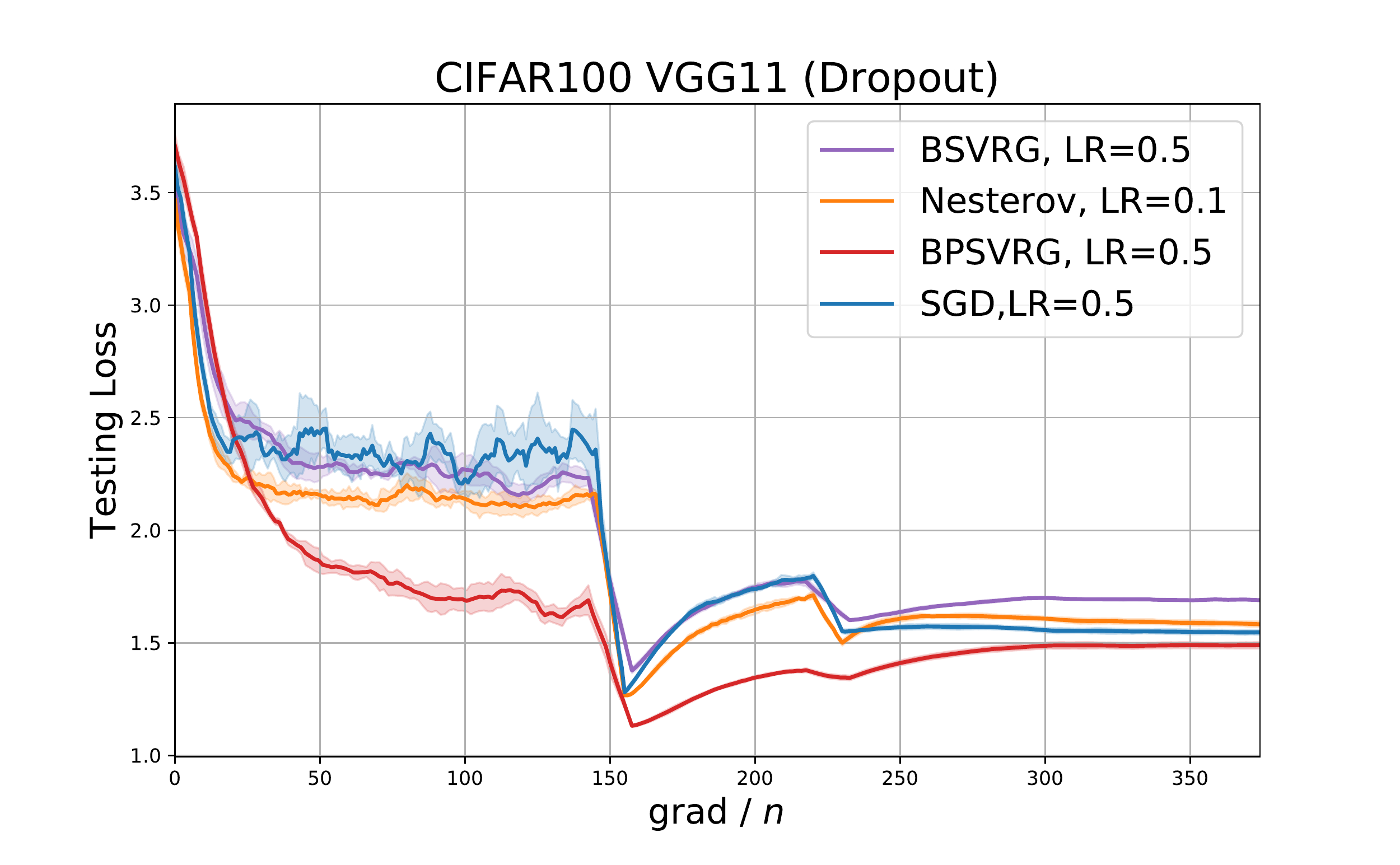}
	\includegraphics[width=0.35\linewidth]{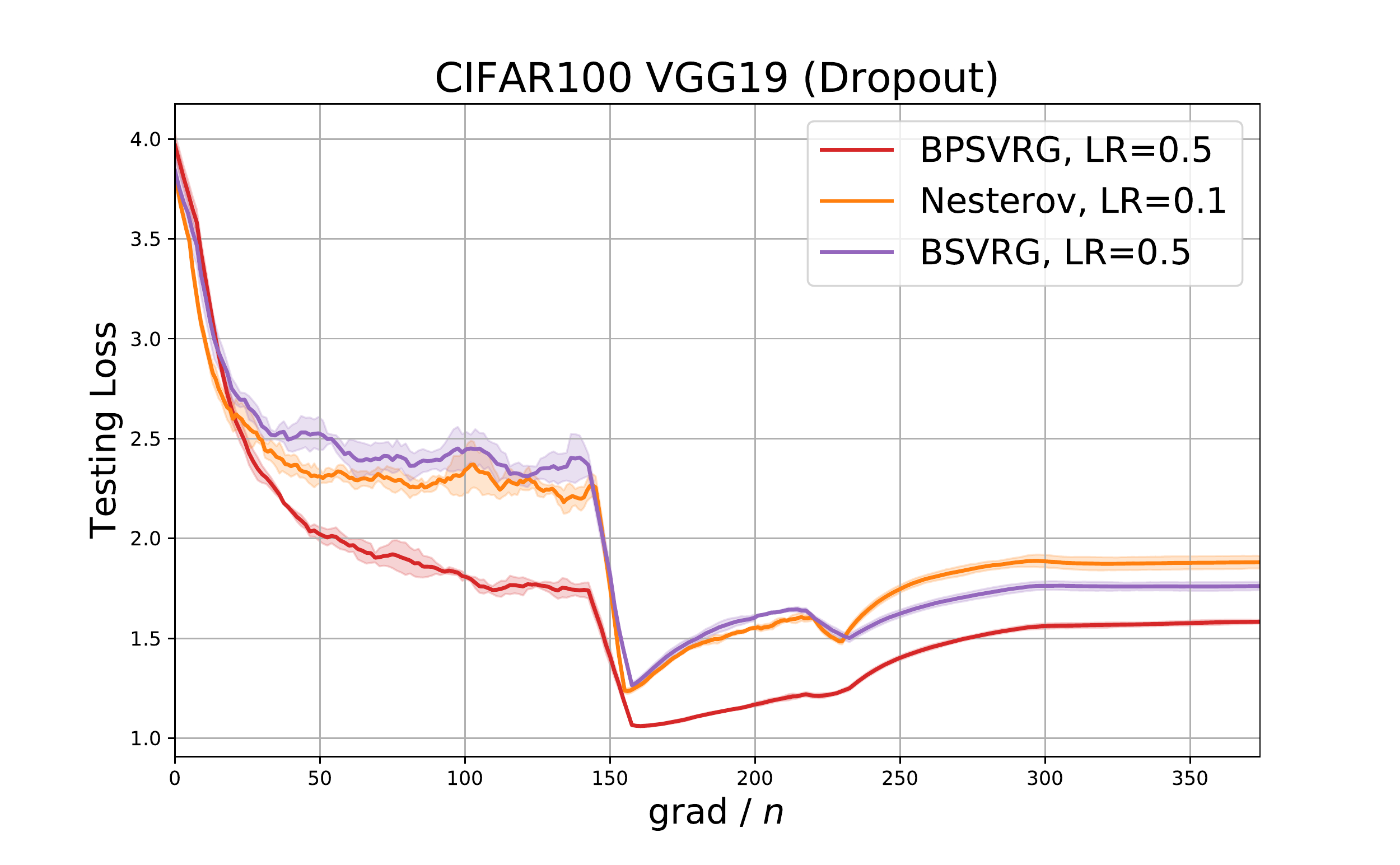}
	\hspace{-0.6cm}
	\includegraphics[width=0.35\linewidth]{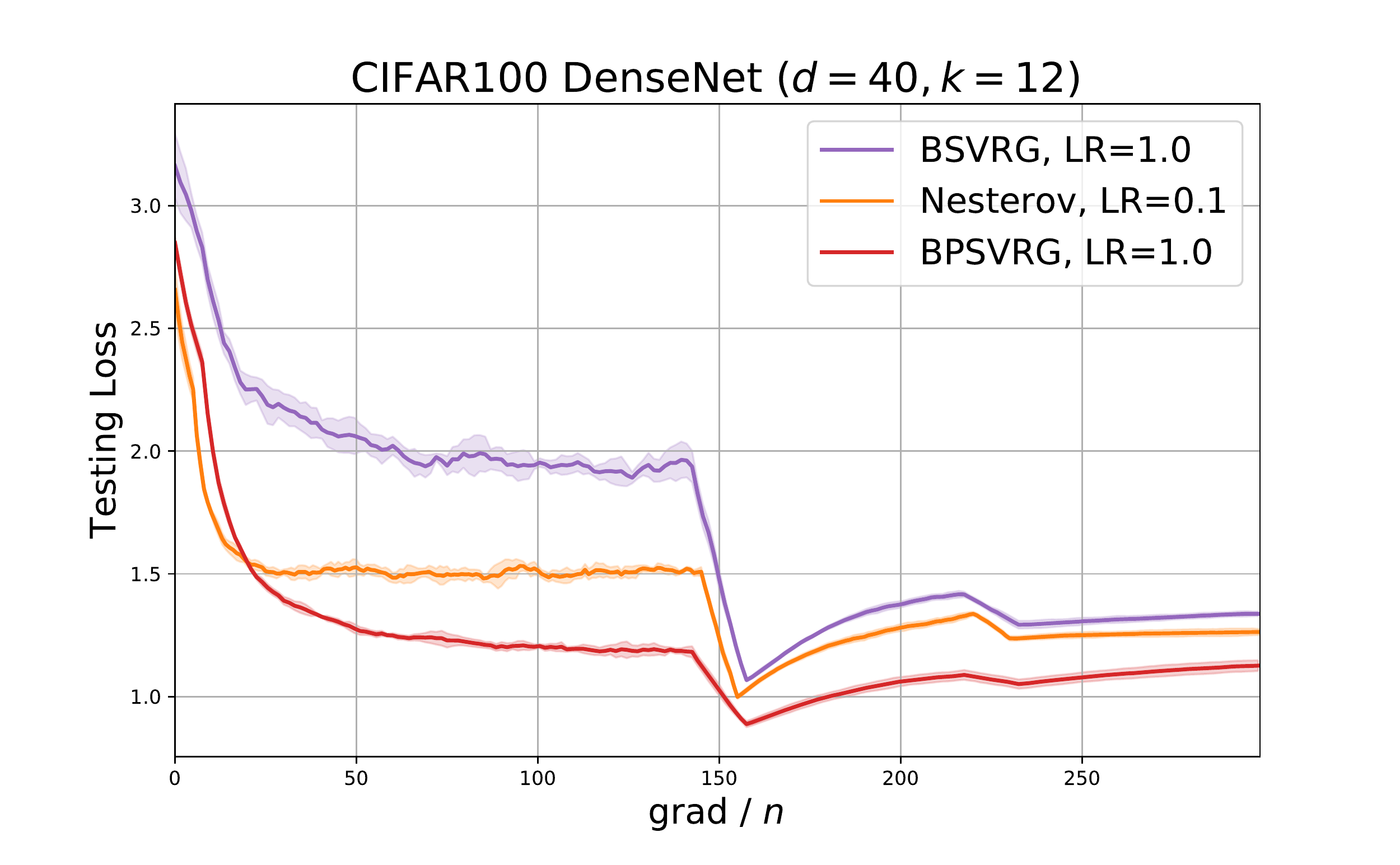}
	\hspace{-0.6cm}
	\includegraphics[width=0.35\linewidth]{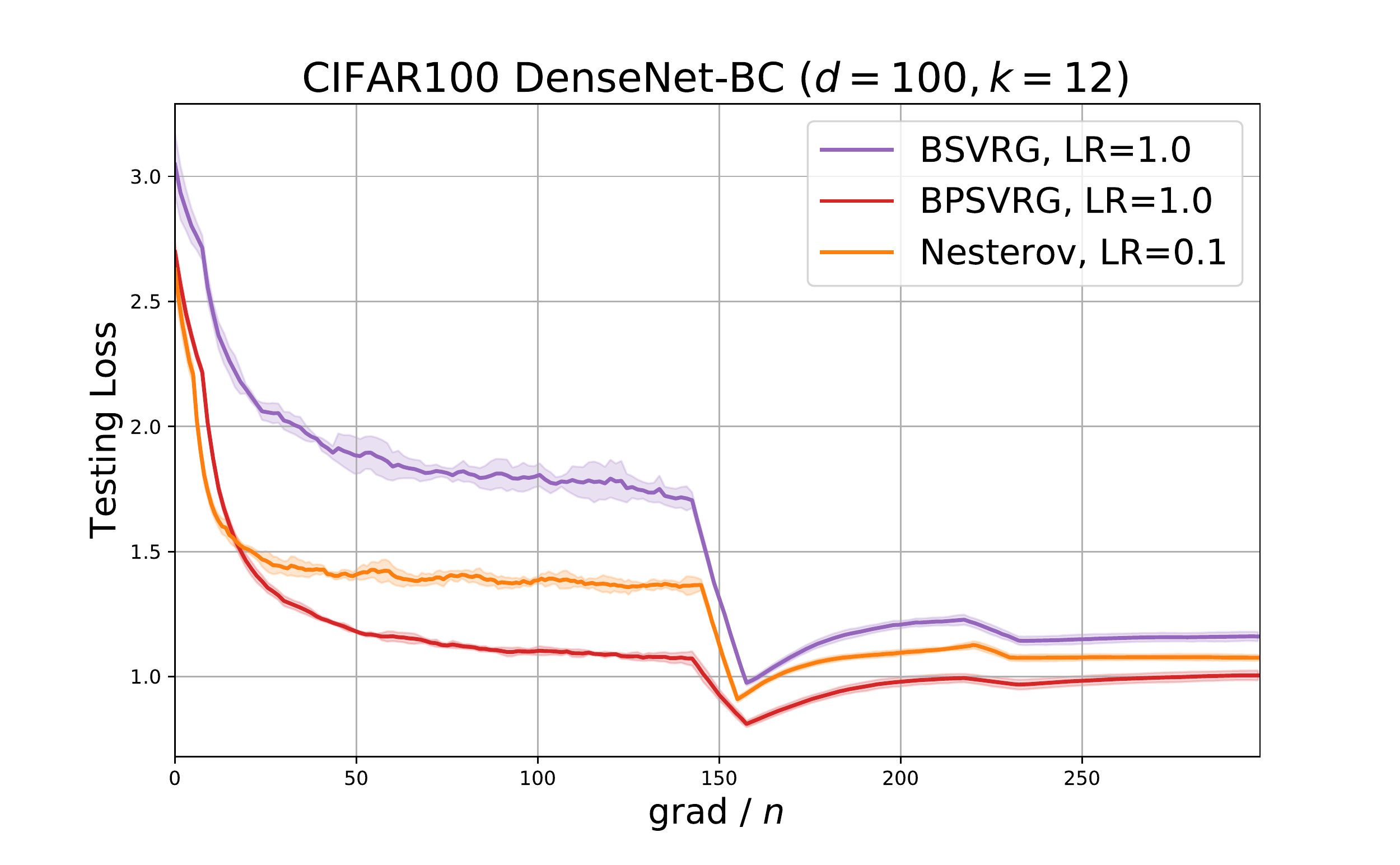}
	\includegraphics[width=0.35\linewidth]{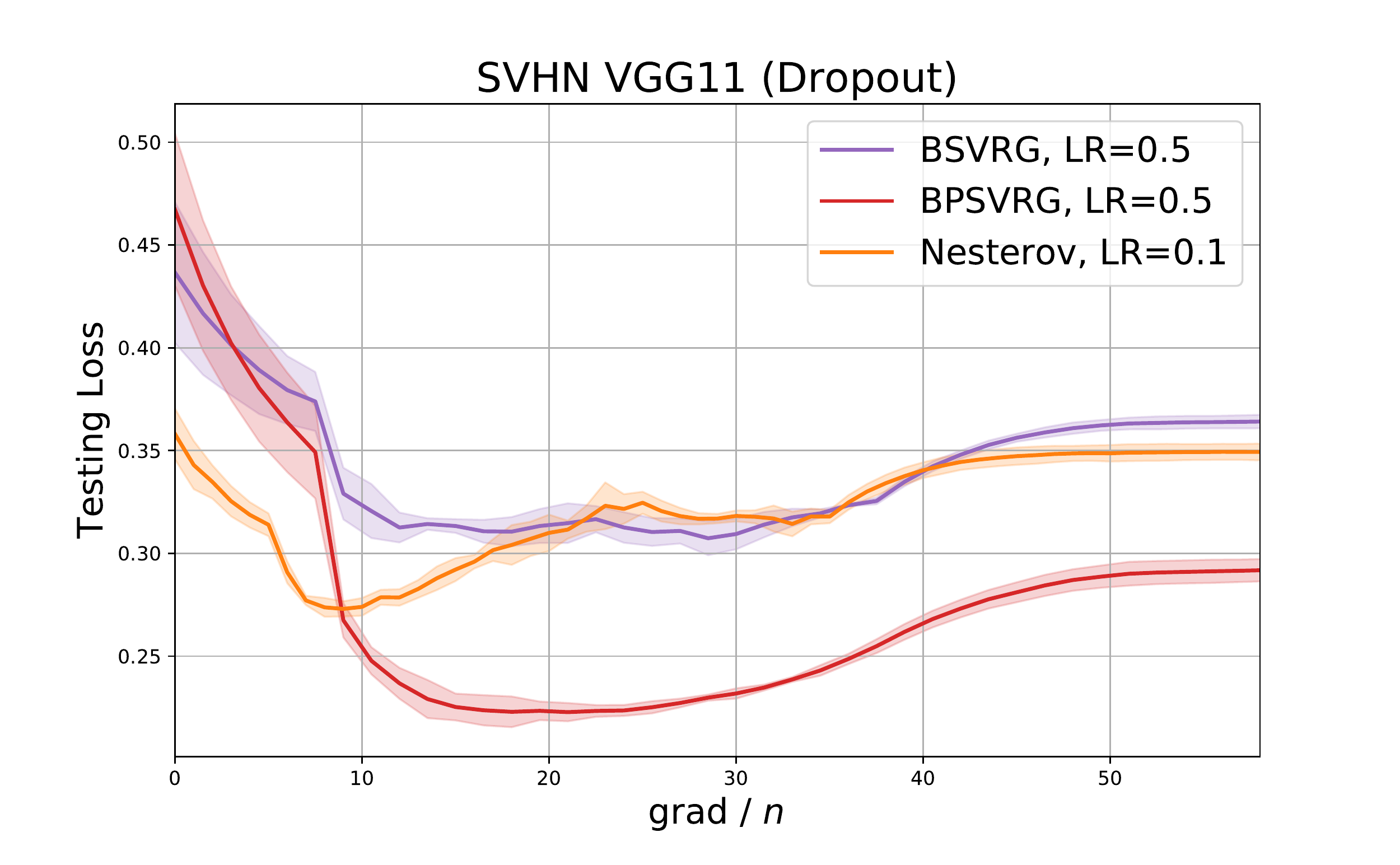}
	\hspace{-0.6cm}
	\includegraphics[width=0.35\linewidth]{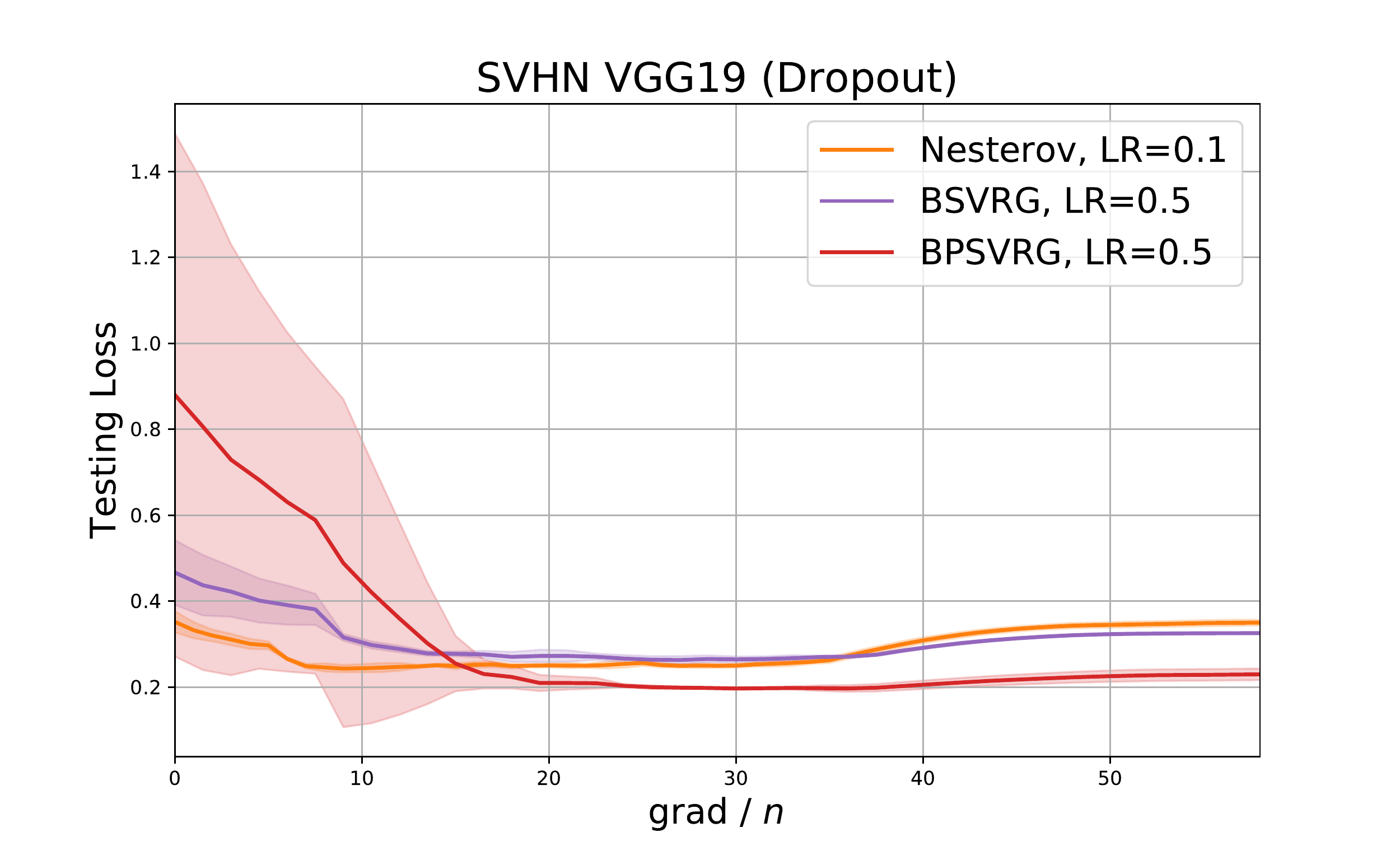}
	\hspace{-0.6cm}
    \includegraphics[width=0.35\linewidth]{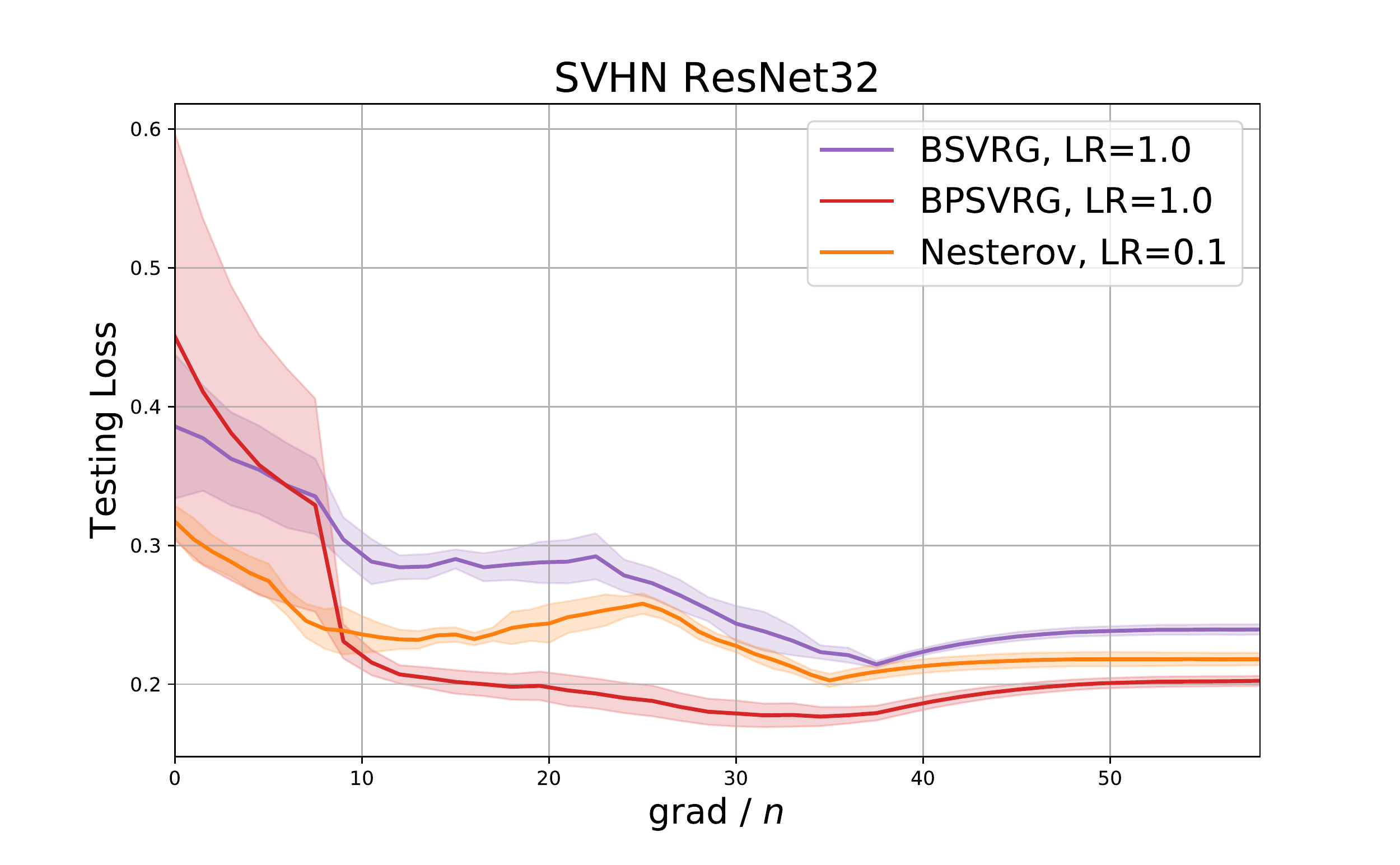}
	\includegraphics[width=0.35\linewidth]{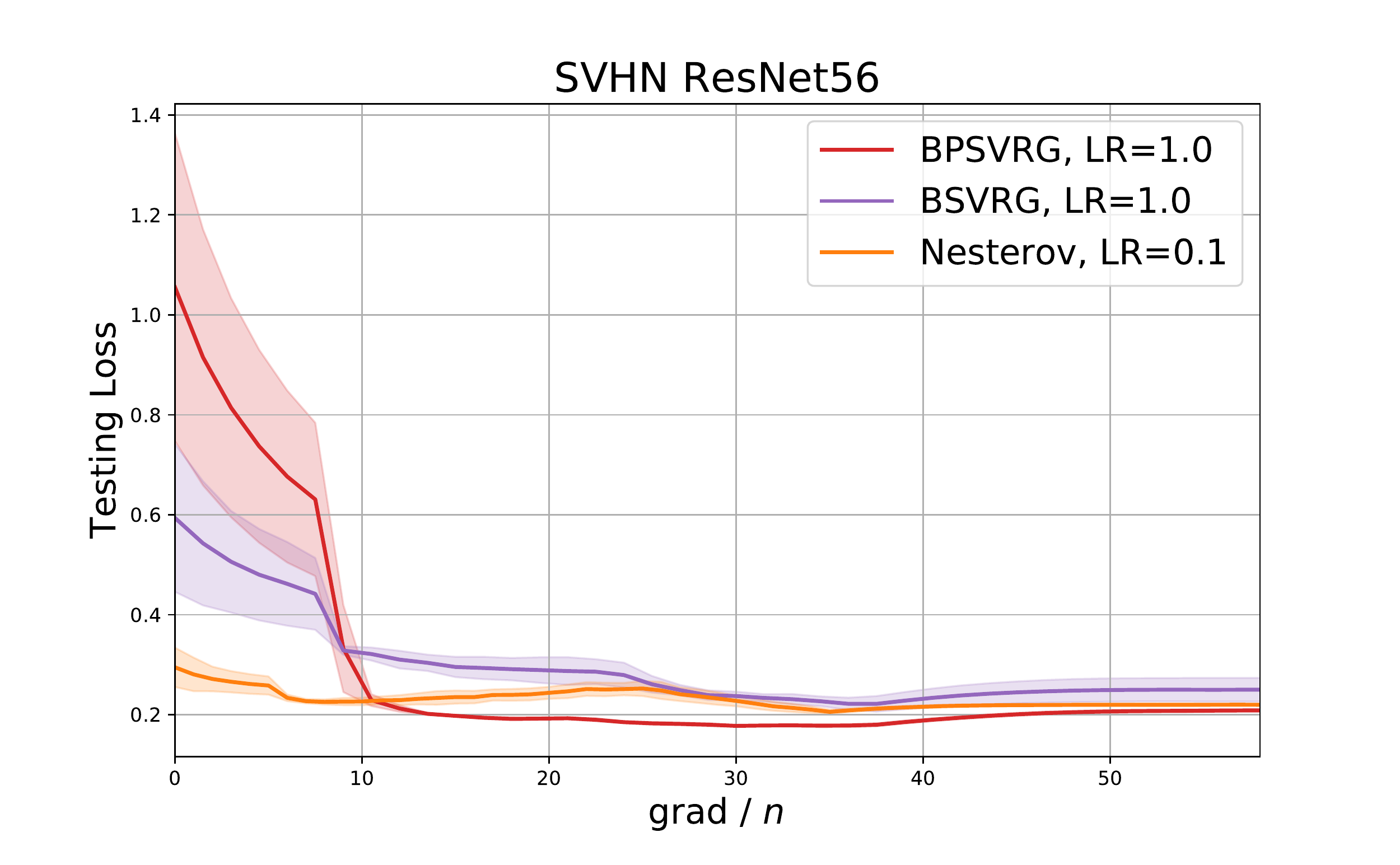}
	\hspace{-0.6cm}
	\includegraphics[width=0.35\linewidth]{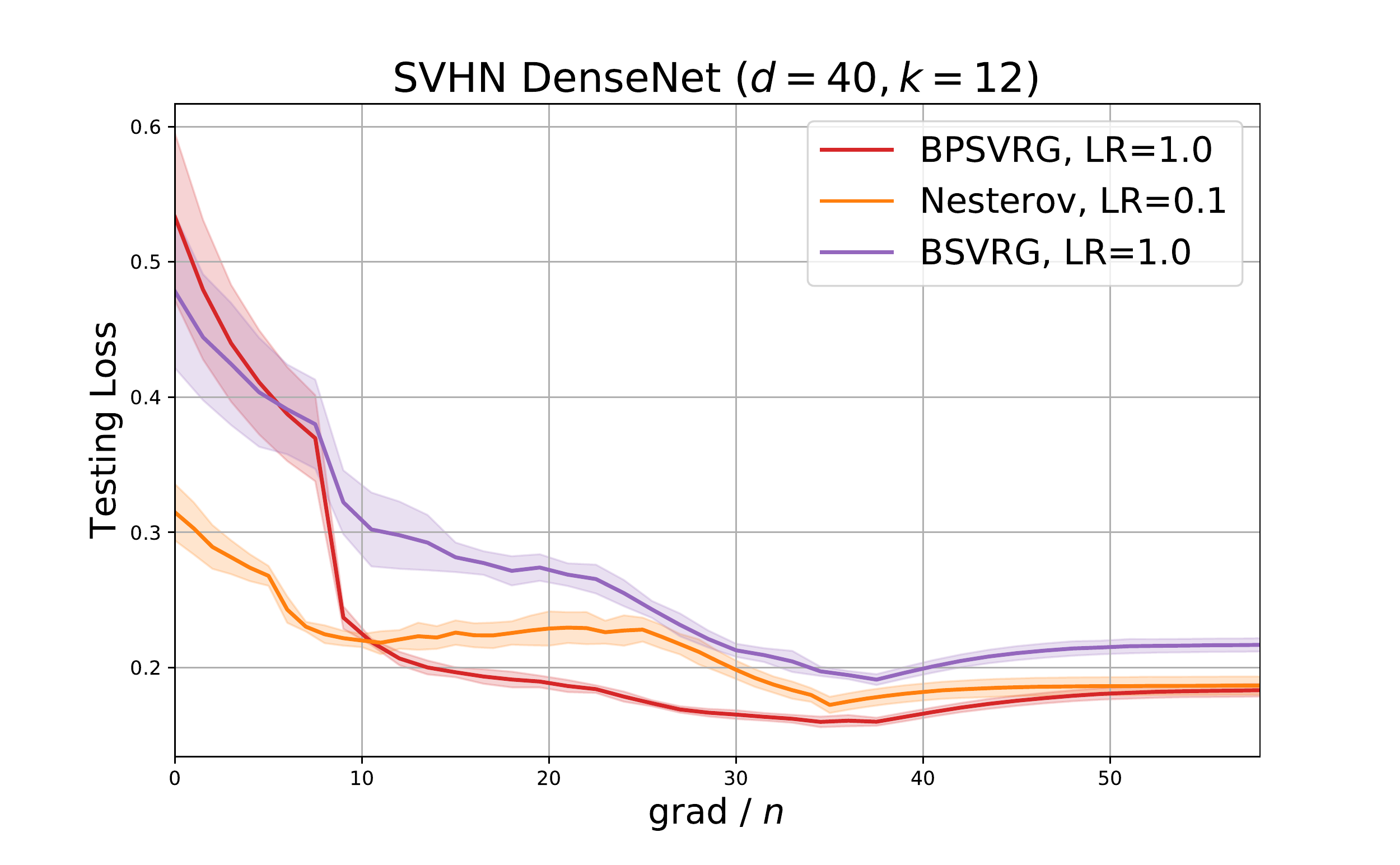}
	\hspace{-0.6cm}
	\includegraphics[width=0.35\linewidth]{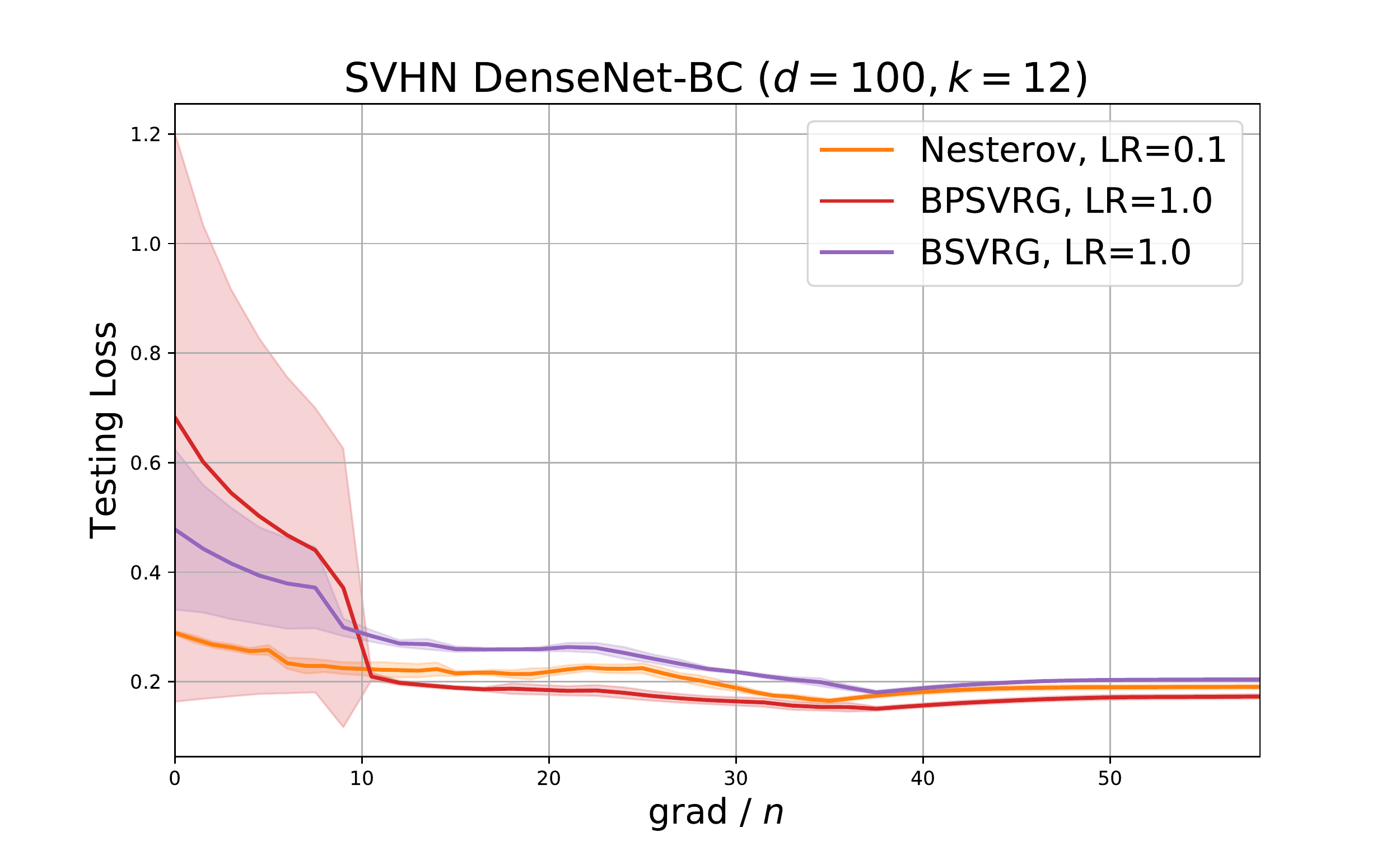}
    \caption{Testing loss (an average $\pm$ standard deviation) on VGG, ResNet, and DenseNet on CIFAR and SVHN. Curves are smoothed with a moving average over 5 points.}
	\label{testloss2}
\end{figure}

Additional testing loss for analysis performed in Section \ref{num-res} are presented in Figure \ref{testloss2}.
As well as the the discovery in Figure \ref{testloss1}, we observe that BP-SVRG offers better performance in all the testing loss figures on CIFAR and SVHN. 
Due to learning rate decay, overfitting may appear and cause increasing regions in testing loss. 
Besides, BPSVRG may need warm-up by some simple optimization methods, SGD to decrease experimental variance in the initial period from our repeated results.

\section{Addition Metrics in Other Networks} \label{supp_var_figure}
We respectively present additional observations of average norm of gradients, loss gap and accuracy gap.
On VGG11, BP-SVRG holds identical conclusions shown in Figure \ref{aveandfg}. 
Furthermore, we run SVRG and PSVRG in shallow networks to confirm our discovery in deep networks. 

\begin{figure}[!ht]
	\centering
    \begin{subfigure}[b]{1\textwidth}
    	\centering
    	\includegraphics[width=0.35\linewidth]{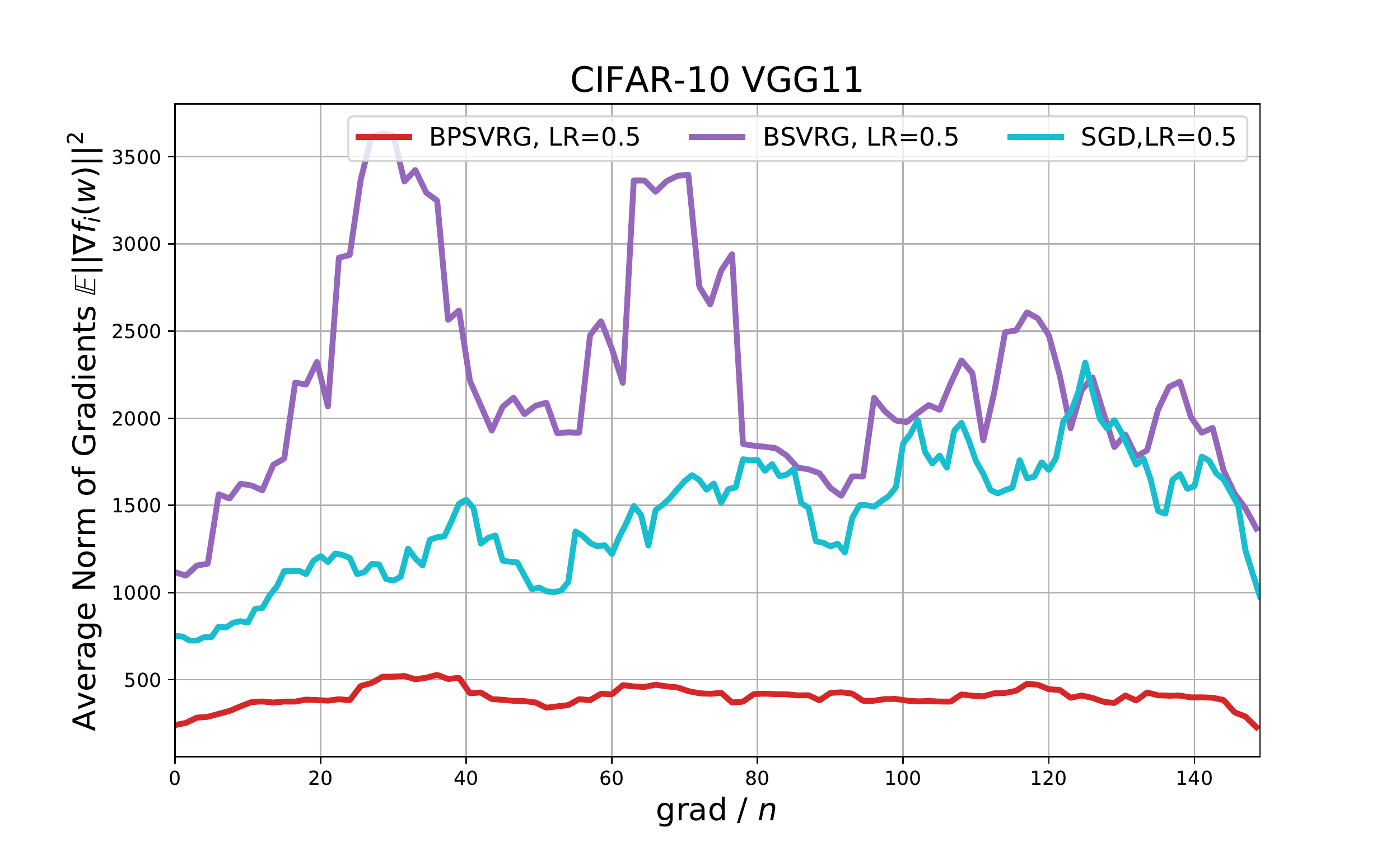}
	    \hspace{-0.6cm}
    	\includegraphics[width=0.35\linewidth]{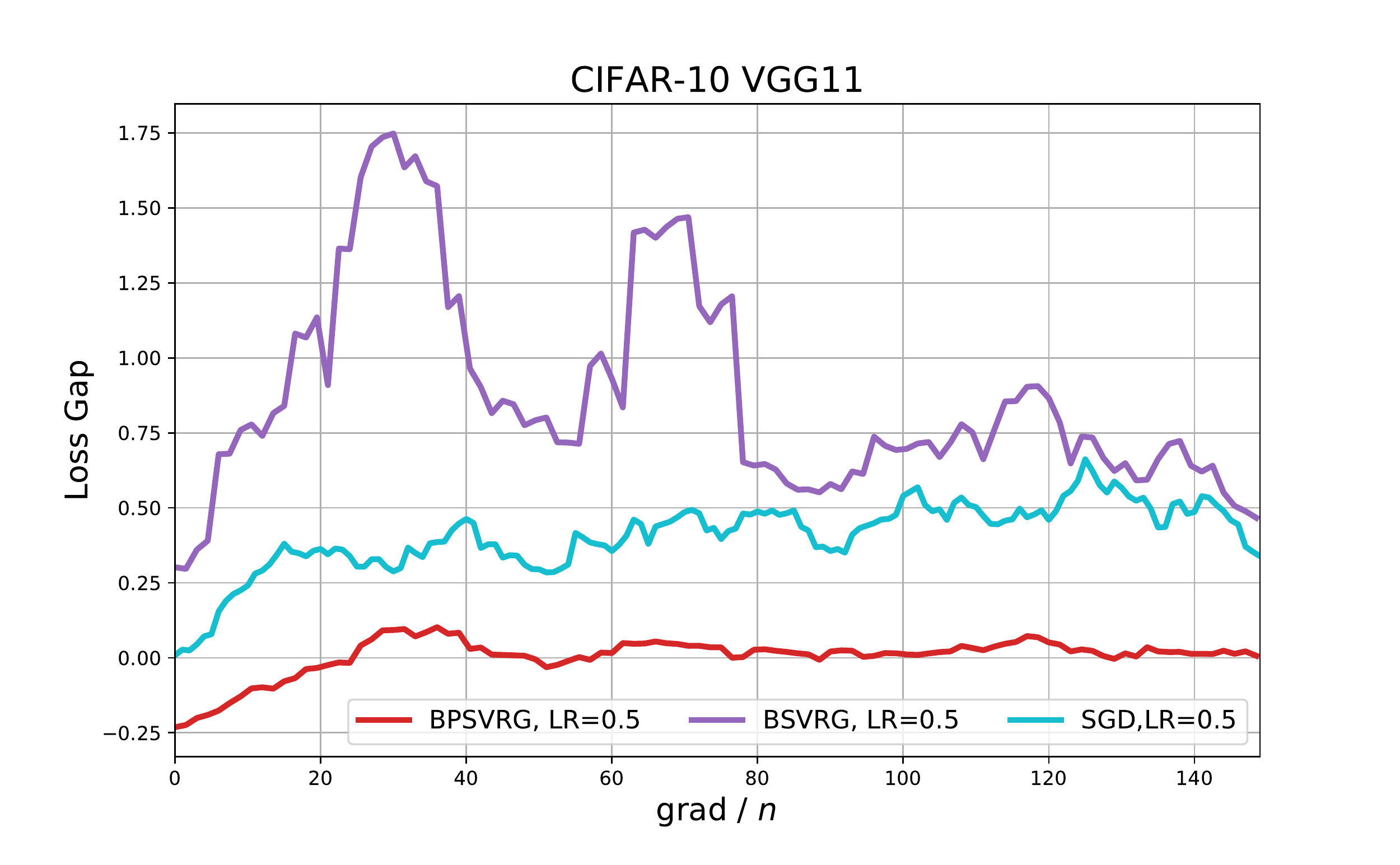}
    	\hspace{-0.6cm}
    	\includegraphics[width=0.35\linewidth]{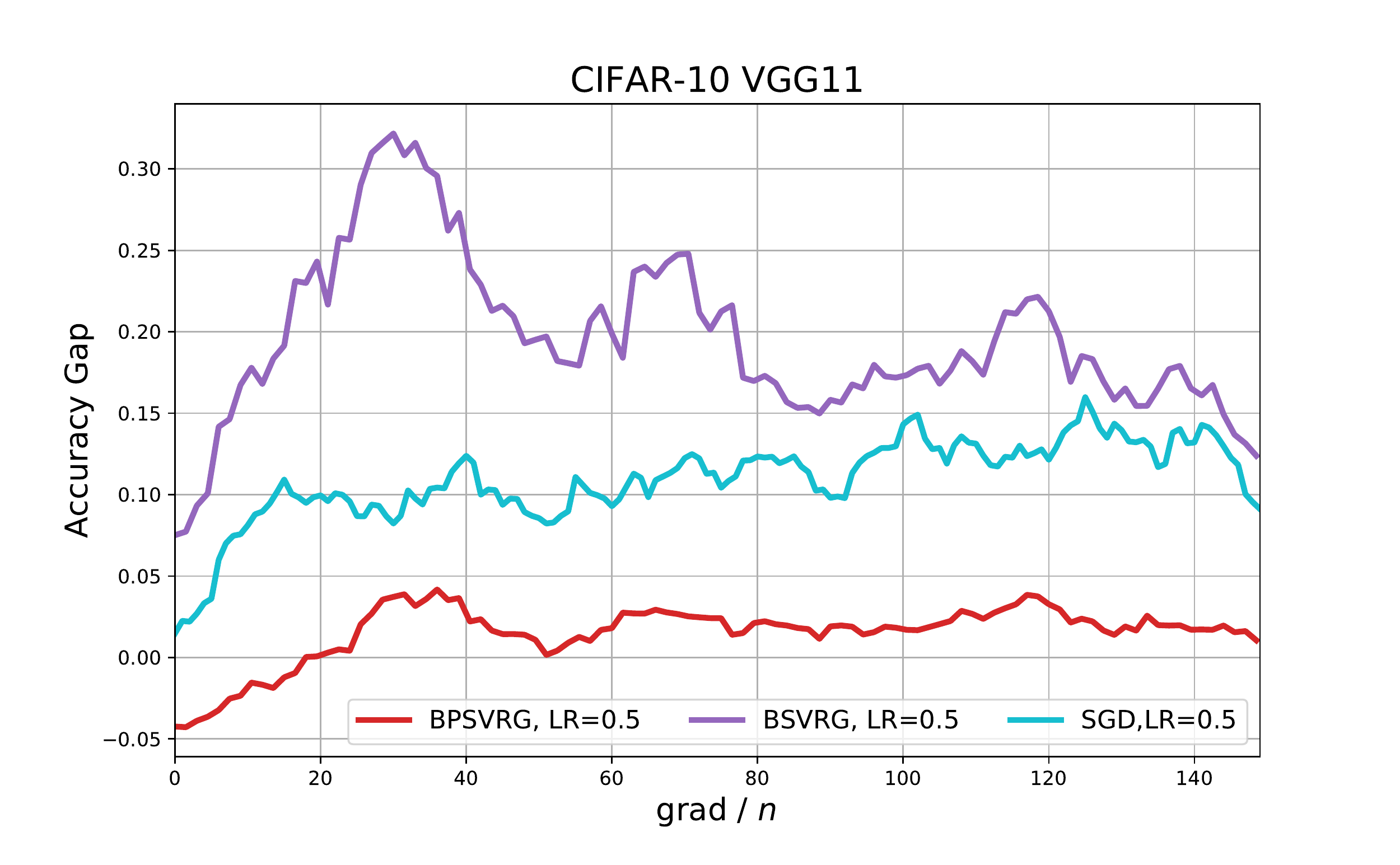}
		\vspace{-0.2cm}
    	\caption{$\mathbb{E}_i\|\nabla f_i(w)\|^2$, Loss Gap, and Accuracy Gap on VGG11}
    \end{subfigure}
    \begin{subfigure}[b]{1\textwidth}
    	\includegraphics[width=0.35\linewidth]{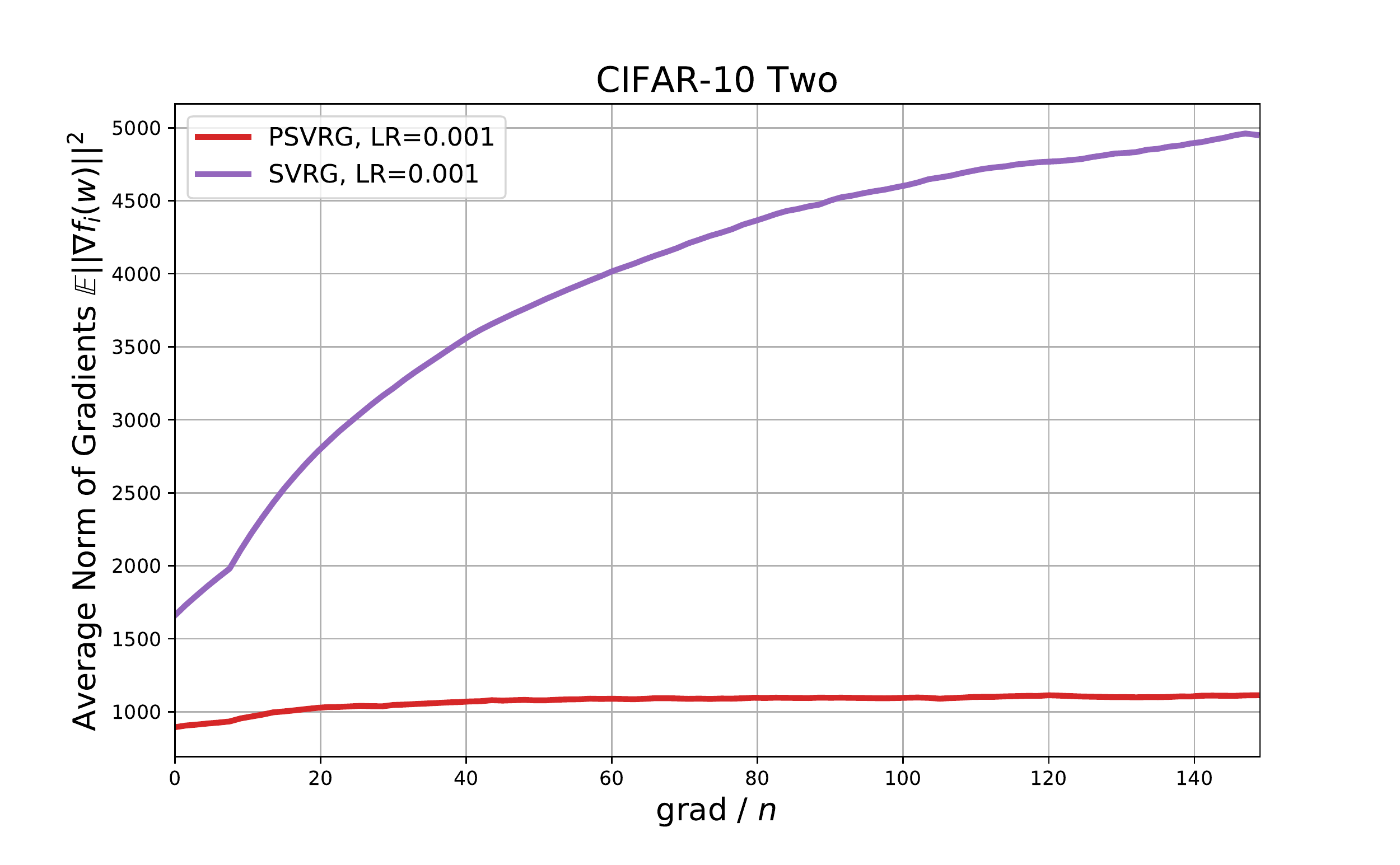}
	    \hspace{-0.6cm}
    	\includegraphics[width=0.35\linewidth]{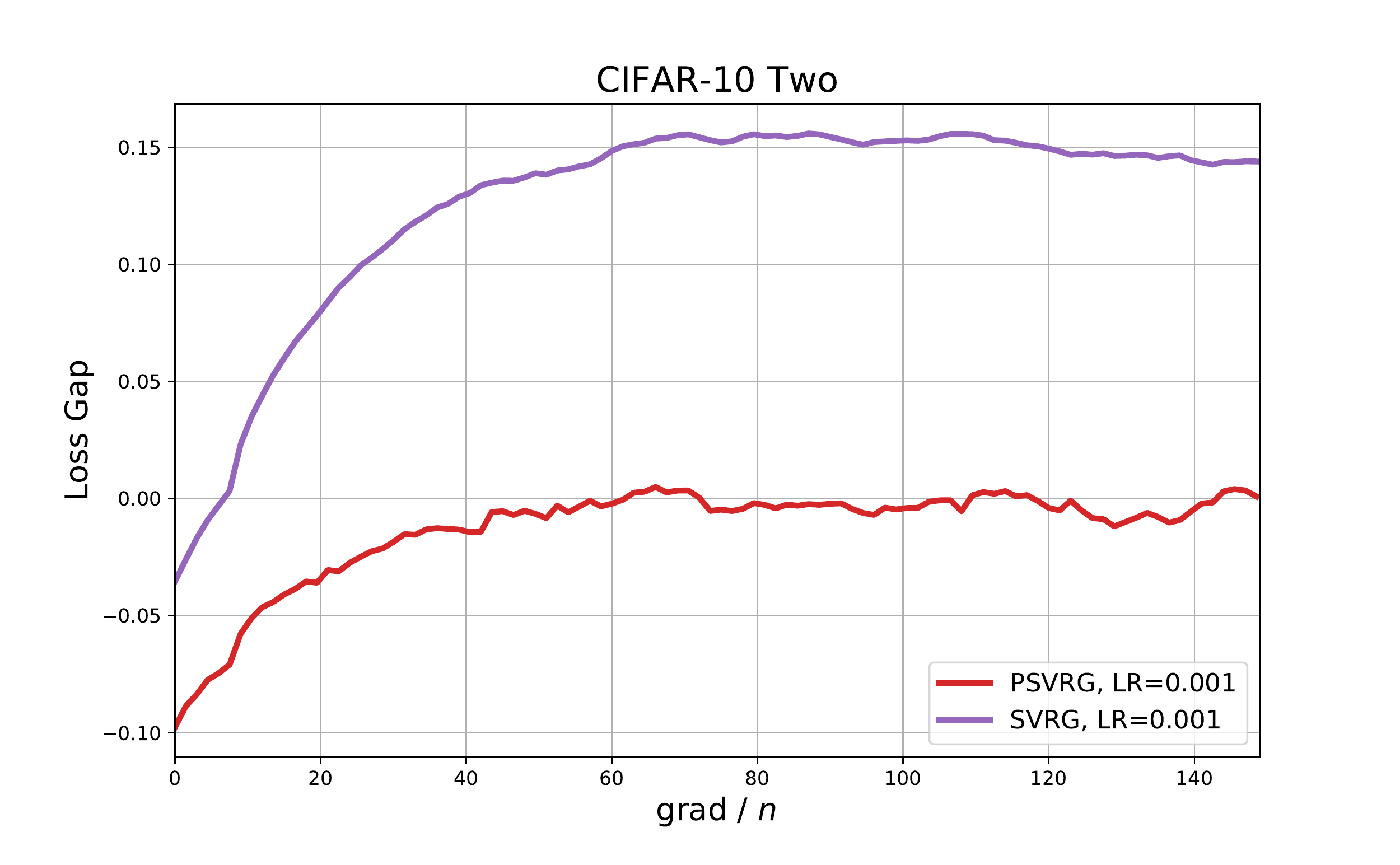}
        \hspace{-0.6cm}
    	\includegraphics[width=0.35\linewidth]{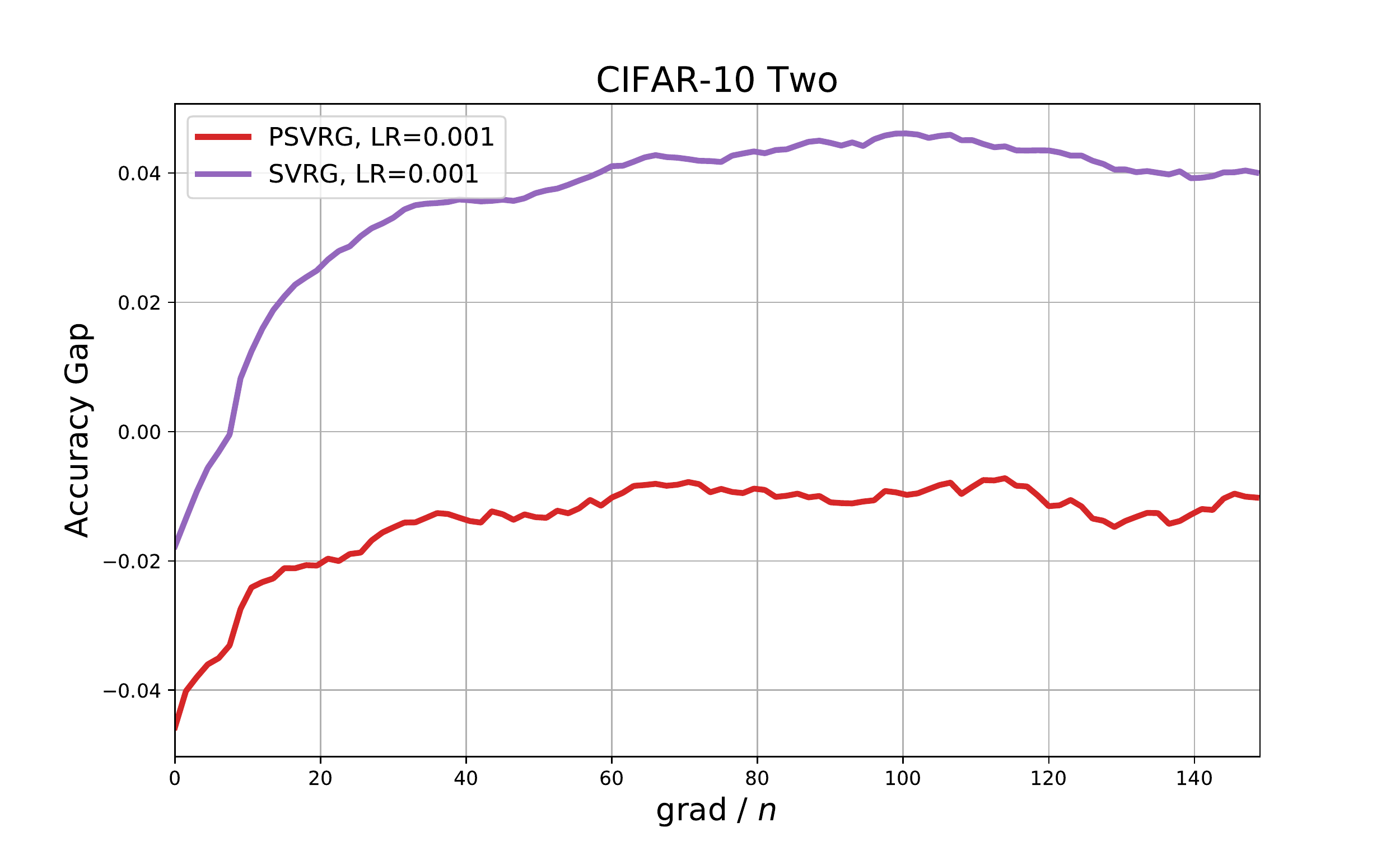}
    	\vspace{-0.2cm}
    	\caption{$\mathbb{E}_i\|\nabla f_i(w)\|^2$, Loss Gap, and Accuracy Gap on two-layer fully-connected NN}
    \end{subfigure}
    % \vspace{-1cm}
	\caption{(a) exhibits the metrics on VGG11, CIFAR10. Experimental settings are same as Section \ref{num-res}. 
	(b) depicts the performance of SVRG and PSVRG in two-layer fully-connected neural network with 100 neurons in each layer and batch size 10. 
	Initial learning rates are shown in the legends and all experiments don't use learning rate decay. Curves are smoothed with a moving average over 5 points.
	Loss gap refers to testing loss minus training loss. Accuracy gap displays the difference between training and testing accuracy. }
	\label{ave_g_and_fg2}
\end{figure}